\documentclass{article}
\usepackage[utf8]{inputenc}
\pdfoutput=1
\usepackage[sc,osf]{mathpazo}
\usepackage[margin=1.1in,letterpaper]{geometry}
\usepackage[sort&compress,numbers]{natbib}
\usepackage[colorlinks=true,citecolor=blue,breaklinks]{hyperref}
\usepackage[hyphenbreaks]{breakurl} 

\makeatletter
\newlength\aftertitskip     \newlength\beforetitskip
\newlength\interauthorskip  \newlength\aftermaketitskip

\def\maketitle{\par
 \begingroup
   \def\thefootnote{\fnsymbol{footnote}}
   \def\@makefnmark{\hbox to 4pt{$^{\@thefnmark}$\hss}}
   \@maketitle \@thanks
 \endgroup
\setcounter{footnote}{0}
 \let\maketitle\relax \let\@maketitle\relax
 \gdef\@thanks{}\gdef\@author{}\gdef\@title{}\let\thanks\relax}

\def\@startauthor{\noindent \normalsize\bf}
\def\@endauthor{}
\def\@starteditor{\noindent \small {\bf Editor:~}}
\def\@endeditor{\normalsize}
\def\@maketitle{\vbox{\hsize\textwidth
 \linewidth\hsize \vskip \beforetitskip
 {\begin{center} \LARGE\@title \par \end{center}} \vskip \aftertitskip
 {\def\and{\unskip\enspace{\rm and}\enspace}%
  \def\addr{\small\it}%
  \def\email{\hfill\small\tt}%
  \def\name{\normalsize\bf}%
  \def\AND{\@endauthor\rm\hss \vskip \interauthorskip \@startauthor}
  \@startauthor \@author \@endauthor}
}}

\makeatother

\pdfoutput=1                    

\usepackage[utf8]{inputenc} 
\usepackage[T1]{fontenc}    
\usepackage{hyperref}       
\usepackage{url}            
\usepackage{booktabs}       
\usepackage{amsfonts}       
\usepackage{nicefrac}       
\usepackage{microtype}      

\usepackage{caption}
\usepackage{graphicx}
\usepackage{amsmath,amsthm,amssymb,bm} 
\usepackage{amsfonts}
\usepackage{mathrsfs}
\usepackage{subfigure}
\usepackage{xspace}
\usepackage{array}
\usepackage{enumerate}
\usepackage{algorithm}
\usepackage{algorithmic}
\usepackage{stmaryrd}
\usepackage{appendix}
\usepackage{wrapfig}
\numberwithin{equation}{section}
\usepackage[colorlinks=true,citecolor=blue]{hyperref}
\usepackage{xcolor}
\usepackage[sort&compress,numbers]{natbib}
\usepackage{appendix}
\usepackage{booktabs}

\theoremstyle{plain}
\newtheorem{theorem}{Theorem}

\newtheorem{lemma}[theorem]{Lemma}

\theoremstyle{definition}
\newtheorem{definition}[theorem]{Definition}

\theoremstyle{remark}

\newcount\Comments  
\usepackage{color}
\definecolor{darkgreen}{rgb}{0,0.5,0}
\definecolor{purple}{rgb}{1,0,1}
\newcommand{\comm}[2]{\ifnum\Comments=1\textcolor{#1}{#2}\fi}

\newcommand{\bp}{\mathbb{P}}

\title{Robust GANs against Dishonest Adversaries}
\author{\name Zhi Xu \email{zhixu@mit.edu}\\
\name Chengtao Li \email{ctli@mit.edu}\\ 
  \name Stefanie Jegelka \email{stefje@csail.mit.edu} \\
  \addr{Massachusetts Institute of Technology}
}

\begin{document}

\maketitle
\begin{abstract}
  Robustness of deep learning models is a property that has recently gained increasing attention. {We explore a notion of robustness for generative adversarial models that is pertinent to their internal interactive structure,} and show that, perhaps surprisingly, the GAN in its original form is not robust. 
  Our notion of robustness relies on a perturbed discriminator, or noisy, adversarial interference with its feedback. We explore, theoretically and empirically, the effect of model and training properties on this robustness. In particular, we show theoretical conditions for robustness {that are supported by empirical evidence. We also test the effect of regularization.}
  Our results suggest variations of GANs that are indeed more robust to noisy attacks and have more stable training behavior, requiring less regularization in general. {Inspired by our theoretical results, we further extend our framework to obtain a class of models related to WGAN, with good empirical performance.}
  Overall, our results suggest a new perspective on understanding and designing GAN models from the viewpoint of {their internal} robustness.
%
\end{abstract}

\section{Introduction}
\label{sec:intro}
In recent years, the adversarial training of generative models (GANs)~\citep{goodfellow2014generative} has received much attention and found numerous applications, including realistic image generation, text to image synthesis, 3D object generation, and video prediction \citep{reed2016generative, 3dgan, vondrick2016generating}. Despite their success, GANs have known training instabilities \citep{goodfellow2016nips}, and many recent works address this issue by either modifying the objective function, the network architecture or training dynamics~\citep{nowozin2016f,goodfellow2016nips,salimans2016improved,arjovsky2017wasserstein,gulrajani2017improved,arjovsky2017towards,huang2017stacked,zhao2016energy,miyato2018spectral,karras2018progressive,durugkar2017generative,bora2018ambientgan:,brock2018large}.

In general, empirical instability in machine learning may be closely related to notions of \emph{robustness}. Robustness has emerged as an important (but often lacking) property of deep learning models. In supervised models, this is exemplified by adversarial examples that fool state-of-the-art, human-level classifiers \citep{szegedy2013intriguing,goodfellow2014explaining}. While there has been a large body of work on robust discriminative models, robust generative models (GANs) are less well studied.
Recent work \citep{bora2018ambientgan:,thekumparampil2018robustness} studies the robustness of GANs to corrupted images (or labels in the conditional case). In this work, we explore an orthogonal route: the success of GANs relies on intricate internal interactions between a generator and a discriminator, which provides another source of vulnerability or instability. A better understanding of the stability of those interactions offers a better understanding of GANs, and may suggest improved, more stable models that are robust to internal perturbations and also train more stably. Hence, in this work, we aim to (1) theoretically characterize notions of robustness that capture internal sensitivities specific to GANs, and (2) identify model properties that affect those sensitivities. 

\begin{figure}[t]
 \centering
   \includegraphics[width=3.5in]{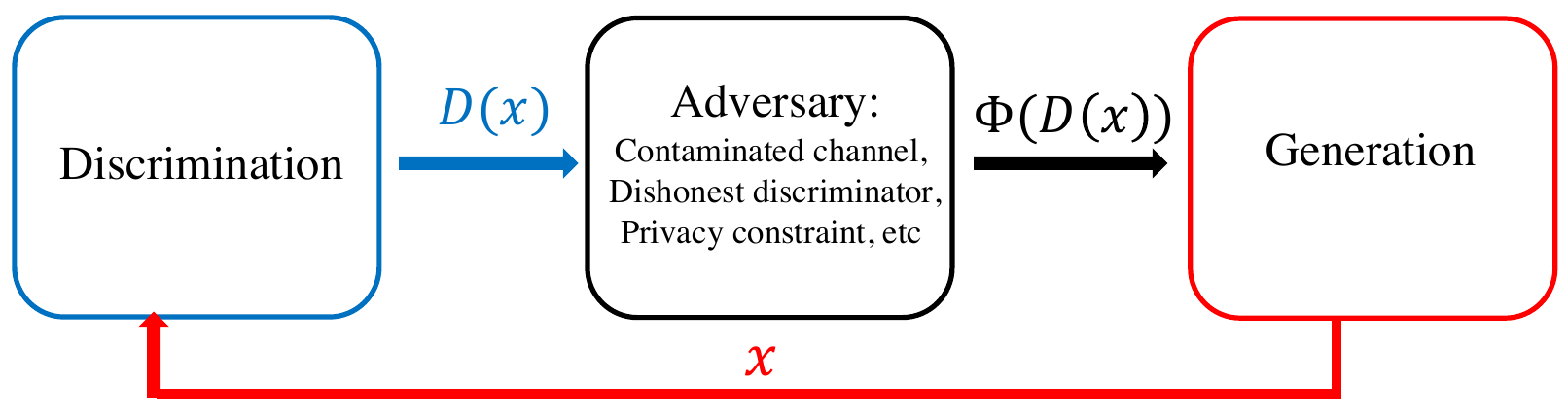}
   \vspace{-0.0in}
   \caption{{GAN with internal perturbation, i.e., an adversary changing the discriminator's feedback.}}\label{fig:gan_adv}
   \vspace{-0.17in}
 \end{figure}

Despite the terminology \emph{generative adversarial networks}, the discriminator in GANs may be viewed as taking on a cooperative, ``teaching'' role, and sharing useful feedback with the generator part \citep{goodfellow2016nips}. In particular, this teacher does not play the same role that an adversary plays in the type of adversarial training that makes discriminative models more stable. Yet, this interaction is key to the learning process. Here, we study the robustness of this interaction by, possibly adversarially, interfering with the interaction, and endowing the teacher with a simultaneous adversarial role. This leads to a new notion of robustness for GANs. Specifically, we perturb the feedback from discriminator to the generator, i.e., with a certain probability, the generator receives a perturbed signal from the discriminator, illustrated in Figure~\ref{fig:gan_adv}. 

Our framework has several interpretations. First, in GANs, the only signal the generator receives about the data is via the discriminator, so, similar to notions of robustness perturbing the data, this perturbation also perturbs the channel from data to the learner. Second, this framework
may be viewed as a {constrained} \emph{dishonest discriminator} (instead of a mere ``teacher''), or as an adversary interfering with the channel between discriminator and generator. This viewpoint relates to ideas in differential privacy \citep{dwork2014algorithmic} where perturbations are used to hide information. E.g., the data may be stored in private databases, and the discriminator teaches but simultaneously masks the data to preserve privacy.

{\bf Contributions.} 
In short, we make the following contributions:
%
(1) We formulate a new notion of robustness for GANs that {naturally} arises from its internal interactive learning dynamics. 
(2) We show that, perhaps surprisingly, the original GAN is not robust in this sense, even to very small perturbations.
(3) We establish general theoretical conditions on the model (objective function) that induce robustness and investigate them empirically.
{(4) Inspired by our theoretical results, we extend our framework and obtain a class of models that relates to and extends Wassterstein GANs, and performs well empirically.}
We will also see that WGAN-like linear losses belong to our robust class, providing a new perspective on their empirical success. 

{\bf Further Related Work.} Most existing work on GANs aims at improving among three directions: formulations~(objective functions), network architectures, and training dynamics. Examples include deriving objective functions based on different divergence measures between probability distributions \citep{nowozin2016f,arjovsky2017wasserstein}; adding multiple discriminators or generators, or an unsupervised channel in GANs \cite{chrysos19}, exploiting that multiple sources may improve stability \citep{durugkar2017generative,tolstikhin2017adagan,nguyen2017dual}; or devising algorithms that better regularize the training process \citep{salimans2016improved,gulrajani2017improved,uehara2016generative,karras2018progressive}. 
Our work
mostly studies the first direction, but
takes a different perspective. In addition, robustness in general has recently gained much attention in machine learning, notably due to the presence of adversarial examples \citep{szegedy2013intriguing,goodfellow2014explaining}. There is a growing body of work on understanding attack and defense mechanisms \citep{fawzi2015analysis,papernot2016distillation,carlini2017adversarial,carlini2017towards,he2017adversarial,tramer2018ensemble,samangouei2018defense,athalye2018obfuscated,xie2019feature,zhang2019theoretically}. Several works \citep{madry2018towards,yang2019me,cohen2019certified} rely on perturbing the input data during training. Distributionally robust optimization \citep{gao17,shafi15,namkoong17} that perturbs the data-generating distribution has also been applied in adversarial training 
\citep{sinha2018certifiable}. At a higher level, studying the perturbation of input data is not new, and many ideas in modern robust machine learning are inherently connected to classical robust statistics \citep{huber1981robust} and robust optimization \citep{ben2009robust,caramanis11}. 
Yet, the actual definition of robustness and the perturbation can be model-dependent and may require new formulations. In particular, to the best of our knowledge, robustness in the context of GANs has not yet been studied formally.  

\section{Failure of GANs for a Simple Adversary}\label{sec:gan_fail}
We begin with an illustrative example: the perhaps surprising observation that the standard GAN can fail even with a rare perturbation of the discriminator. The standard GAN's objective function is:
%
\begin{align}\label{vanilla_gan}
\begin{split}
 \min_G \max_D V(G,D)&= \min_G \max_D \{\mathbb{E}_{x\sim \bp_{\text{data}}} [\log D(x)]  + \mathbb{E}_{z\sim \bp_z } [\log (1- D(G(z)))] \},
\end{split}
\end{align}
where $D: \mathbb{R}^d \rightarrow [0, 1]$ is a discriminator that maps a sample to the probability that it comes from the true data distribution $\bp_{\text{data}}$, and $G: \mathbb{R}^l \rightarrow \mathbb{R}^d$ is the generator that maps a noise vector $z \in \mathbb{R}^l$, drawn from a simple distribution $\bp_z$, to the data space. This in turn defines an implicit distribution $\bp_G$ for $G$'s generated data.
It can be shown that, when fixing $G$, the optimal discriminator $D$ is given by $D^*_G(x)=\frac{\bp_{data}(x)}{\bp_{data}(x)+\bp_G(x)}$~\citep{goodfellow2014generative}. 
The generator then essentially seeks to minimize 
$V(D^*_G, G) = -\log(4) + 2\times \text{JSD}(\bp_{\text{data}}\|\bp_G),$
and the optimal generator would give $\bp_G = \bp_{\text{data}}$.

Existing work implicitly assumes that during training, the discriminator is \emph{honest}, i.e., it always gives ``true feedback'' to the generator about how likely it deems the generated sample to come from $\bp_{\text{data}}$. What happens if this no longer holds true? E.g., there could be channel contamination, adversarial interventions, or constraints such as privacy that prevent the discriminator from releasing precise feedback.
%
Formally, we treat such \emph{dishonest feedbacks} as applying a transformation $\Phi(\cdot):[0,1]\to[0,1]$ to the discriminator's outputs so that the generator receives $\Phi(D(x))$. One may view $\Phi$ as an adversary that encodes 
what the generator actually receives. \emph{Not} knowing about the existence of such an adversary, the generator regards $\Phi(D(x))$ as honest feedback.

Ideally, we desire a robust GAN model: if $\Phi$ does not alter the original outputs $D(x)$ too much, the model should still be able to learn the data distribution. Is this true for the standard GAN? Let us consider a simple flipping adversary defined as follows:
\begin{align}\label{flipping_adversary}
\Phi(D(x)) = \left\{\begin{array}{ll}
1-D(x) & \;\text{with probability $p$}\\
D(x) & \;\text{otherwise.}
\end{array}\right.\end{align}
That is, with error probability $p$, the feedback is flipped to be $1-D(x)$. 
Note that we assume the signal from $G$ to $D$ to be always correct, i.e., $D$ always receives the original real and generated data. As such, the optimal discriminator is still $D_G^*$. With the flipping adversary, the minimization problem for $G$ then becomes
\begin{align}\label{vanilla_g_dishonest}
\begin{split}
\min_{G}\: p & \big\{\mathbb{E}_{x\sim \bp_{\text{data}}}[\log(1- D_G^*(x))] + \mathbb{E}_{x\sim \bp_G}[\log D_G^*(x)]\big\}+\\
&(1-p)\big\{\mathbb{E}_{x\sim \bp_{\text{data}}}[\log D_G^*(x)] + \mathbb{E}_{x\sim \bp_G}[\log(1-D_G^*(x))]\big\}.
\end{split}
\end{align}

\begin{lemma}\label{claim:vanilla_gan}
Given the optimal discriminator $D^*_G$, the minimization of the objective (\ref{vanilla_g_dishonest}) becomes
\begin{align}\label{lemma1:obj}
\min_{G}\quad  2 \times JSD(\bp_{\text{data}}||\bp_G)-\log(4)
-p \left\{KL(\bp_{\text{data}}||\bp_G) + KL(\bp_G||\bp_{\text{data}})\right\}. 
\end{align}
Furthermore, {for every $p>0$, the optimal $\bp_G$ can be arbitrarily far from $\bp_{\text{data}}$ in terms of the KL-divergence.}
\end{lemma}
{For an intuitive understanding, note that for any $p>0$, if $\bp_G=\bp_{data}$, then the objective function in~\eqref{lemma1:obj} becomes 0. But, because of the term $-p \left\{KL(\bp_{\text{data}}||\bp_G) +\right.$ $\left. KL(\bp_G||\bp_{\text{data}})\right\}$, the objective function can be much smaller; it can be $-\infty$. To see this, note that the Jensen-Shannon divergence is bounded, but the KL-divergence is not. Any $\bp_G$ that has a disjoint support from $\bp_{data}$ can make the above term $-\infty$. For example, a learned distribution $\bp_G$ that only concentrates on a particular mode of $\bp_{data}$ with no coverage on the other modes is optimal, achieving $-\infty$ for the objective function in~\eqref{lemma1:obj}. Such a behavior of mode collapse is highly undesirable for the generator.}

Essentially, 
Lemma~\ref{claim:vanilla_gan} establishes that even for very small perturbations, the GAN is not robust: even if the discriminator is almost always honest, it fails to extract sufficient information from the data. This observation raises the question  whether it is possible to construct a robust GAN, and, if so, what types of adversaries other than the flipping adversary it can defend against. Next, we formally define families of adversarial attacks and corresponding conditions for robustness.


\section{GAN with Dishonest Discriminators}
Motivated by the simple flipping adversary, we next consider a broader class of adversaries. This will lead to a more general framework and notion of robustness.

\subsection{Dishonest Discriminators}
We formalize \emph{dishonest discriminator feedbacks} as post-processing the original outputs $D(x)$ (or corresponding gradients) by an adversary. Since $D(x)$ is typically viewed as the probability of the data coming from the true distribution, the transformed feedback should still lie in the range $[0,1]$. More explicitly, we use a differentiable transformation function $\psi:[0,1]\to[0,1]$ as \emph{perturbation} or \emph{dishonest function}.
The flipping adversary \eqref{flipping_adversary} consists of two perturbations: $\psi_1(y)=1-y$ and $\psi_2(y)=y$. An adversary can combine several such transformations into a more complex attack:
%
\begin{definition}[Adversary]\label{def:attack}
Let $\Psi$ be a set of perturbations. An \emph{adversary} $\Phi$ with respect to $\Psi$ is a probability distribution over finitely many perturbations, $\psi_1,\psi_2\ldots\psi_L\in\Psi$. Denote by $p_i$ the probability the adversary assigns to the $i$th perturbation $\psi_i$. Given input $y\in[0,1]$, the adversary outputs $\Phi(y) = \psi_i(y)$ with probability $p_i$. 
\end{definition}
Definition \ref{def:attack} generalizes the flipping adversary defined in \eqref{flipping_adversary} to more powerful and flexible attacks.
In general, we do not expect to be able to construct GANs that are robust against all possible adversaries -- imagine the adversary always replaces the signal with random noise. Instead, we will assume that most of the time the feedback is honest.
\begin{definition}[Mostly Honest Adversary] \label{def:mostly_honest}
An adversary $\Phi$ is \emph{mostly honest} if the probability it assigns to the function $\psi(y)=y$ is larger than 0.5.
\end{definition}
We will refer to a GAN as \emph{robust} if it learns the data distribution with a mostly honest adversary. Intuitively, it seems reasonable that a mostly honest adversary should retain sufficient signal to learn, if the learning is not too sensitive to perturbations. Yet, with this definition of robustness, the standard GAN is not robust.

\subsection{GAN Formulation with Adversaries}
Before adding an adversary, we revisit the GAN objective in Eq.~\eqref{vanilla_gan}. The $\log$ function in the objective was suggested because of its nice information-theoretic interpretation. Recent variants such as the Wasserstein GAN \citep{arjovsky2017wasserstein} replace the $\log$ with other functions. In a unified framework, one could think of the GAN objectives as:
  \begin{align}
    \max_{D}\: \mathbb{E}_{x\sim \bp_{\text{data}}}[f_D( D(x))]+\mathbb{E}_{z\sim \bp_z}[f_D(1-D(G(z)))],\label{gen_att_dis}\\
    \min_{G}\:\mathbb{E}_{x\sim \bp_{\text{data}}}[f_G( D(x))] + \mathbb{E}_{z\sim \bp_z}[f_G(1-D(G(z)))].\label{gen_att_gen}
  \end{align}
In the standard GAN, $f_D(\cdot)=f_G(\cdot)=\log(\cdot)$.
  %
%
In presence of an adversary $\Phi$, the generator receives transformed feedback $\Phi(D(x))$ instead of $D(x)$, as shown in Figure \ref{fig:gan_adv}. {However, this information is not known to the generator. In other words, \emph{without}} knowing the existence of such an adversary, the generator treats $\Phi(D(x))$ as if it is $D(x)$. The generator's objective then becomes
 \begin{align} \label{gen_att_gen2}
   \min_{G}\:&\mathbb{E}_{x\sim \bp_{\text{data}},\Phi}[f_G( \Phi(D(x)))]
   + \mathbb{E}_{z\sim \bp_z,\Phi}[f_G(1-\Phi(D(G(z))))]\\
%
 \label{gen_att_gen3}
 \equiv \min_{G}\:&\sum_{i=1}^Lp_i\Big( \mathbb{E}_{x\sim \bp_{\text{data}}}[f_G( \psi_i(D(x)))] + \mathbb{E}_{z\sim \bp_z}[f_G(1-\psi_i(D(G(z))))]\Big).
 \end{align}
In summary, with an adversary $\Phi$, the discriminator's objective \eqref{gen_att_dis} remains unchanged, because the adversary does not affect what the discriminator receives. In contrast, the generator's objective becomes Eq.~\eqref{gen_att_gen3}.

\section{Robustness against Perturbed {Feedbacks}}\label{sec:robust}
We are now ready to study conditions on $f_D$ and $f_G$ that imply robustness. When designing these functions, we need to keep three aspects in mind: (1) The objective \eqref{gen_att_dis} of the discriminator aims to maximize the probability that the discriminator can distinguish true data from fake data; (2) the objective \eqref{gen_att_gen3} of the generator aims to minimize the probability that the discriminator recognizes the generated data as fake; (3) when there is no adversary or the adversary is mostly honest,  the optimal generator $G^*$ 
should be able to learn the true data distribution, i.e., $\bp_{G^*}=\bp_{data}$.


The first two criteria are easily met by choosing $f_G$ and $f_D$ to be monotonically increasing. For robustness, we already saw that the $\log$ function is not suitable. To construct robust models, we will need the class $\mathcal{H}$ of odd functions around $0.5$ with support $[0,1]$:
%
\begin{equation*}
\mathcal{H}\triangleq\big\{f(\cdot):f \textrm{ is strictly increasing and differentiable}
\textrm{ in $[0,1]$, and }f(\theta)=-f(1-\theta), \forall\: \theta\in[0,1]\big\}.
\end{equation*}
Lemma~\ref{lemma2} characterizes the optimal discriminator when $f_D\in\mathcal{H}$.
\begin{lemma}\label{lemma2}
Suppose that $f_D\in\mathcal{H}$. For a fixed $G$, the optimal $D$ that maximizes the objective \eqref{gen_att_dis} is
\begin{equation}\label{opt_h}
D^*_G(x)=\left\{ \begin{array}{rcl}
1, & \mbox{if }
& \bp_{\text{data}}(x)>\bp_G(x), \\ 0, & \mbox{if } & \bp_{\text{data}}(x)<\bp_G(x), \\
\emph{$[0,1]$}, & \mbox{if} 
& \bp_{\text{data}}(x)=\bp_G(x),
\end{array}\right.
\end{equation} 
where the notation $D^*_G(x)=[0,1]$ means that $D^*_G(x)$ can be any scalar in the interval $[0,1]$.
\end{lemma}
In the following, we construct two GAN frameworks that are robust under mostly honest adversaries. 
The first framework retains the $\log$ function for the discriminator and chooses a function from $\mathcal{H}$ for the generator: 

{\noindent\bf Framework 1:} $f_D=\log(\cdot)$ and $f_G\in\mathcal{H}$.

Theorem~\ref{thm:stronger_1} establishes the robustness of Framework 1 under mild conditions on the perturbations $\psi$:
\begin{theorem}\label{thm:stronger_1}
Suppose that $f_D(\cdot)=\log(\cdot)$ and $f_G\in\mathcal{H}$. Let $\Psi$ be the set of perturbations $\psi:[0,1]\rightarrow[0,1]$ that satisfy {\bf either} one of the following:
\begin{enumerate}
  \item $\psi(\theta)$ is non-decreasing in $[0,1]$ and $\psi(\frac{1}{2})=\frac{1}{2}$;
  \item $\psi(\theta)$ is non-increasing in $[0,1]$, $\psi(\frac{1}{2})=\frac{1}{2}$, and
  \begin{equation*}
\left\{ \begin{array}{rcl}
\psi(\theta)+\theta \geq 1, & \mbox{for }
& \theta\in(\frac{1}{2},1], \\ 
\psi(\theta)+\theta \leq 1, & \mbox{for }
& \theta\in[0,\frac{1}{2}).
\end{array}\right.
\end{equation*}
\end{enumerate}
Then, for any mostly honest adversary $\Phi$ with respect to $\Psi$, given the optimal discriminator $D^*_G$, the optimal generator $G^*$ satisfies $\bp_{G^*}(x)=\bp_{\text{data}}(x)$. 
\end{theorem}

Unlike Framework 1, the second framework we identify uses functions from $\mathcal{H}$ for both the discriminator and the generator. Such a choice leads to a stronger robustness guarantee against mostly honest adversaries, without conditions on the perturbations. 

{\noindent \bf Framework 2:} $f_D\in\mathcal{H}$ and $f_G\in\mathcal{H}$.

\begin{theorem}\label{thm:stronger_2}
Suppose that $f_D\in\mathcal{H}$ and $f_G\in\mathcal{H}$. Let $\Psi$ be the set of all possible perturbations. Then, for any mostly honest adversary $\Phi$ with respect to $\Psi$, given the optimal discriminator $D_G^*$,
the optimal generator satisfies $\bp_{G^*}(x)=\bp_{\text{data}}(x)$.
\end{theorem}

{\emph{Proof of Theorems \ref{thm:stronger_1} and \ref{thm:stronger_2} (Sketch):} Since the adversary is mostly honest, the probability it assigns to the function $\psi(\theta)=\theta$ is larger than 0.5. Without loss of generality, denote by $\psi_1$ the previous function, i.e., $\psi_1(\theta)=\theta$. Both frameworks use a function $f_G$ from the class $\mathcal{H}$ for the generator. By the properties of $\mathcal{H}$, one can show, by rearranging the terms, that the generator's objective~\eqref{gen_att_gen3} can be rewritten as: 
\begin{equation*}
\quad\min_G\quad V_1+V_2,
\end{equation*}
where
\begin{align*}
V_1\triangleq&\Big(p_1-\sum_{i=2}^Lp_i\Big)\Big(\mathbb{E}_{x\sim \bp_{data}}\Big[f_G\big(D(x)\big)\Big]-\mathbb{E}_{x\sim \bp_G}\Big[f_G\big(D(x)\big)\Big]\Big),\\
V_2\triangleq&\sum_{i=2}^Lp_i\Big(\mathbb{E}_{x\sim \bp_{data}}\Big[f_G\big(\psi_i(D(x))\big)+f_G\big(D(x)\big)\Big]-\mathbb{E}_{x\sim \bp_G}\Big[f_G\big(\psi_i(D(x))\big)+f_G\big(D(x)\big)\Big]\Big).
\end{align*}
It is immediate that if $\bp_G=\bp_{data}$, then $V_1=V_2=0$. Now, if we can show that $V_1+V_2$ is greater than $0$ for any $\bp_G\neq\bp_{data}$, the two theorems will be established. This amounts to show that for both frameworks, the following two claims hold:
\begin{enumerate}
  \item If $\bp_{data}\neq \bp_G$, then $V_1>0$.
  \item If $\bp_{data}\neq \bp_G$, then $V_2\geq0$.
\end{enumerate}
The two claims can be proved by considering the different optimal discriminator for each framework. Note that the fact that the adversary $\Phi$ is mostly honest guarantees that the term $(p_1-\sum_{i=2}^Lp_i)$ in $V_1$ is positive. Hence, to establish the first claim, we only need to show that the second term in $V_1$ is positive if $\bp_{G}\neq\bp_{data}$. For the second claim, the terms in $V_2$ involve different perturbations $\psi_i$. This is why Theorem \ref{thm:stronger_1} requires some mild conditions on $\psi$. Under Framework 1, one can show that the second claim holds if those conditions are satisfied. However, for Framework 2, the second claim can be proved without additional conditions. See Appendix A for the details.$\hfill\blacksquare$ }

Theorems \ref{thm:stronger_1} and \ref{thm:stronger_2} show that the flipping adversary is just a special case that our GAN frameworks can defend against. In fact, the theorems provide a stronger robustness guarantee that holds across a variety of mostly honest adversaries.
  The second framework is significantly stronger than the first. In particular, Framework 2 requires no additional conditions on the perturbations. As long as the adversary is mostly honest, robustness is guaranteed. 
  Hence, for robust GANs, it is desirable to use functions in $\mathcal{H}$ instead of the logarithm. 

With $f_D(\theta)=f_G(\theta)=\theta-0.5$, the resulting model not only belongs to Framework 2, the loss functions are also closely related to the well-known Wasserstein GAN. Both models use linear functions in the objectives. The differences are: (1) the last layer of the discriminator in WGAN is linear and hence, the outputs are unnormalized raw scores instead of probabilities; (2) to minimize the dual form of the Wasserstein metric, the discriminator in WGAN is restricted to be a $1-$Lipschitz function.
  Nevertheless, this observation 
  may give further support for their empirical performance, besides the interpretation of using Wasserstein distance.  


\section{Empirical Results: Robustness}\label{sec:empirical_robustness}

\begin{figure}[htp]
\centering
\subfigure[]{
    \label{fig:objective}
    \includegraphics[width=0.465\textwidth]{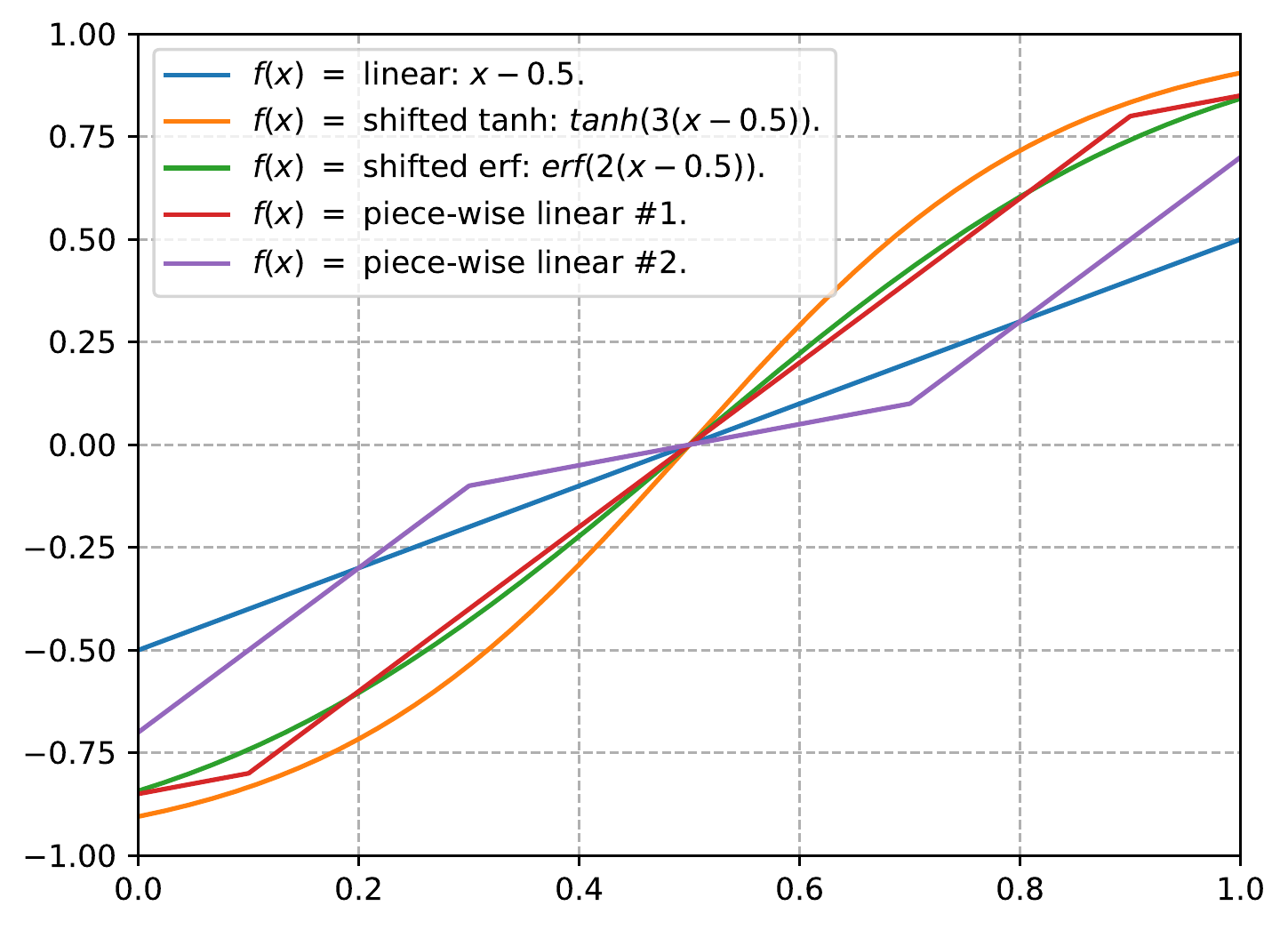}
}
\subfigure[]{
    \includegraphics[width=0.485\textwidth]{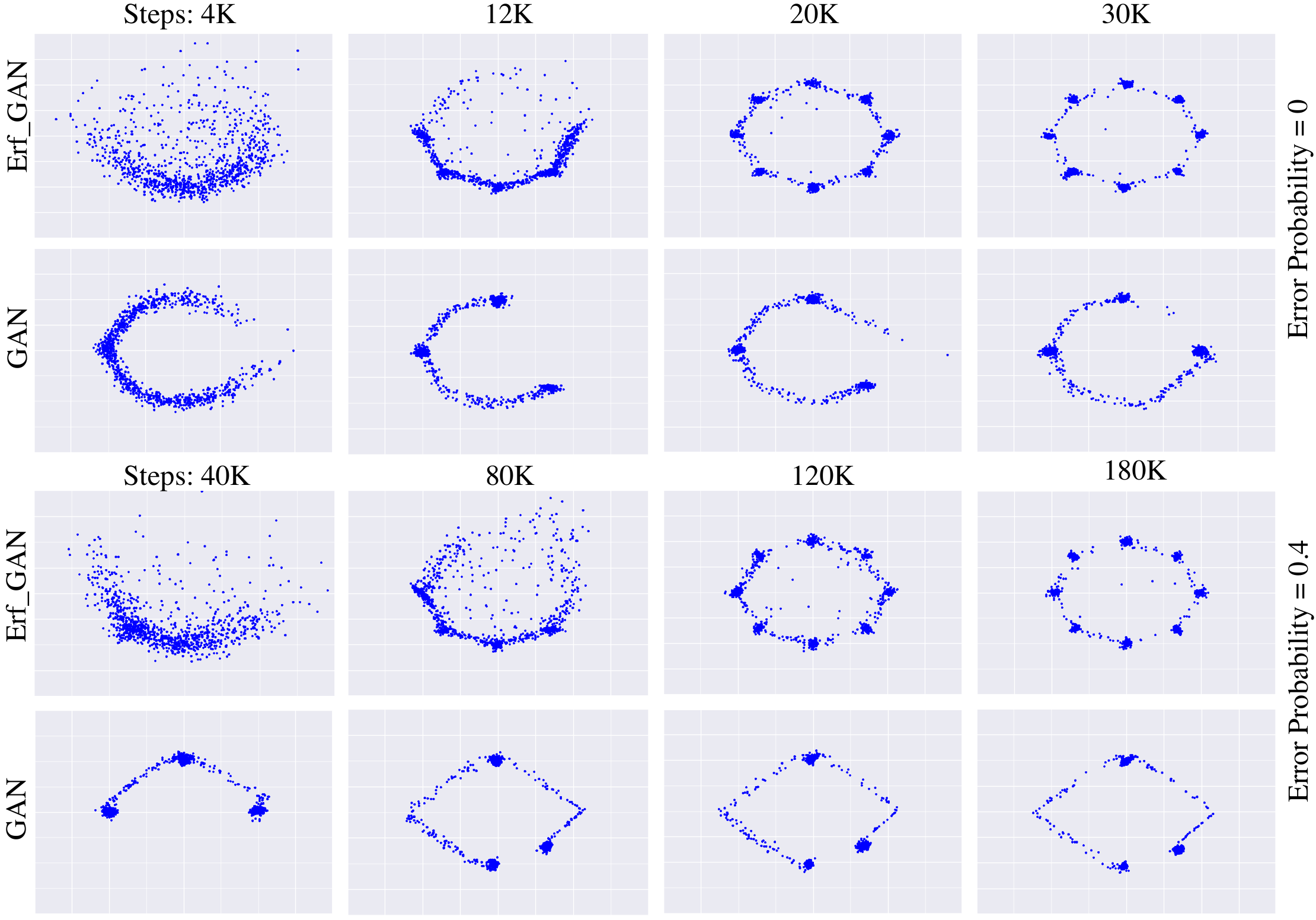} 
\label{fig:gaussian_example}
}
\vspace{-0.3cm}
\caption{(a) Objective functions for robust GANs. (b) Learning processes 
   with flipping adversaries.}
\vspace{-0.3cm}
\end{figure}

To probe our theoretical results in practice, we empirically evaluate the robustness of the models in Section~\ref{sec:robust}. 
Following the convention in most of the GAN literature, we use a zero-sum game formulation, i.e., $f_D=f_G=f$ (Framework 2). We test different mostly honest adversaries, and robustness is guaranteed by Theorem \ref{thm:stronger_2}. Figure~\ref{fig:objective} displays the functions in $\mathcal{H}$ that we investigate. These functions are chosen to have different gradients in different locations, e.g., constant, relatively smaller, or larger gradients around the midpoint. We will refer to the GANs with $f_D, f_G \in \mathcal{H}$ (i.e., Figure~\ref{fig:objective}) as robust GANs, and to the standard GAN with $f_D=f_G=\log(\cdot)$ simply as GAN. Details and additional experiments may be found in Appendix \ref{app:exp_robustness}.

%
%
%

{\bf Regularization and other factors.} 
In practice, apart from the objective function, factors such as the training algorithm and data also influence the outcome of learning. For example, clipping large weights or, in general, regularizing the Lipschitz constant of the discriminator \citep{qi2017loss,uehara2016generative,gulrajani2017improved,miyato2018spectral}, appear to stabilize the overall training process.
In particular, any modifications that result in more averaging and slower adoption of information from single training data points would be expected to make the GAN more robust. Hence, a fair evaluation of our robust objective functions should take these into consideration.
To this end, in our experiments, we also test the effect of clipping as a representative for such regularizing mechanisms, 
and its interplay with the objective function. Clipping was employed in the original Wasserstein GAN \citep{arjovsky2017wasserstein} and, due to the connections between our framework and WGAN mentioned before, 
it is a natural representative choice for regularization. 


\subsection{Synthetic Data: Mixture of Gaussians}\label{sec:exp_gaussian}

We begin with the common illustrative toy problem of a mixture of eight two-dimensional Gaussians. 
Both $G$ and $D$ are fully connected networks. We alternatively train $G$ and $D$, and clip the weights of $D$ with a maximum absolute value of $0.1$. Here, we apply the simple flipping adversary \eqref{flipping_adversary} with different error probabilities $p$. Figure \ref{fig:gaussian_example} shows some typical results for one of the robust models and GAN, for $p=0$ and $p=0.4$.
Indeed, as opposed to the GAN, the robust GAN reliably learns all modes, even with a fairly high $p=0.4$. The figure illustrates an additional intuition: with higher noise $p$, learning indeed becomes more challenging, and the generator needs more iterations to learn. Figure \ref{fig:supp:gaussian_step} in Appendix \ref{app:exp_robustness} confirms this is generally the case.

\begin{figure*}[h]
 \centering
   \includegraphics[width=5.4in]{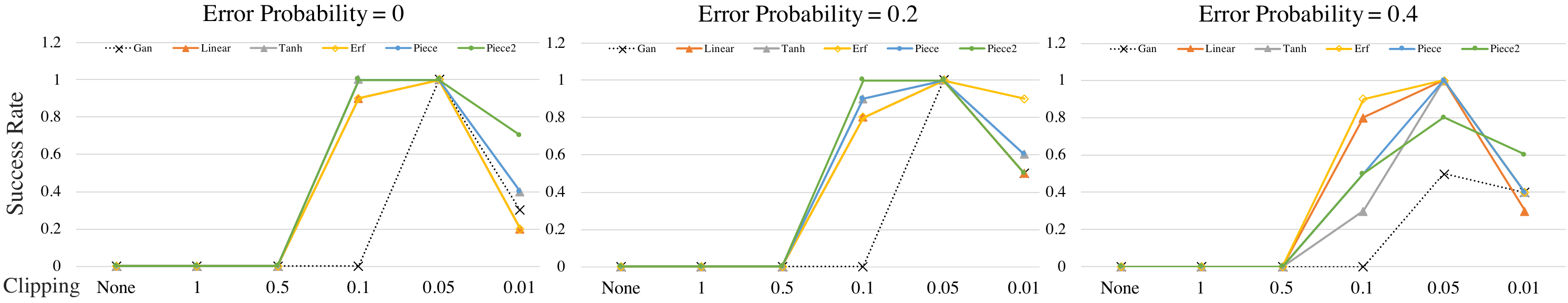}
   \caption{Success rates and average number of learned modes for various models and clipping values. 
   }\label{fig:guassian_1}
   \vspace{-0.3cm}
 \end{figure*}

{\bf Clipping.} 
To probe the effect of regularization, we next vary the clipping threshold. Figure \ref{fig:guassian_1} shows the success rate for different clipping thresholds, averaged over 10 runs. Here, a success is defined as correctly learning \emph{all} the 8 modes (average number of modes are shown in Appendix \ref{app:exp_robustness}). To better visualize the effect of clipping, $D$ is intentionally made more powerful by having significantly more hidden neurons (4x more hidden neurons for each layer). 


Figure~\ref{fig:guassian_1} offers several observations:
(1) The training algorithm indeed affects the results. A too powerful disriminator (small threshold) generally impairs learning \citep{arjovsky2017wasserstein,miyato2018spectral}; very small thresholds limit the capacity of $D$ too much.
(2) However, consistently, if a robust GAN can learn the distribution without noise ($p=0$), then it also learns the distribution with noise, confirming its robustness. In general, the robust GANs work across a wider range of clipping thresholds, i.e., are less sensitive to parameter choice of clipping. These observations support our theoretical analysis in Sections~\ref{sec:gan_fail} and \ref{sec:robust}.


An interesting phenomenon to note is that in some cases, clipping may increase the empirical robustness of the standard GAN, although it is still more sensitive than the robust models (threshold 0.05 and $p=0.4$). This phenomenon is orthogonal to our theoretical results, which focus on the models (objective functions); here, the algorithm aids empirical stability via a stronger regularization. 
This points to the important role of training algorithms in practice 
\citep{salimans2016improved,gulrajani2017improved,uehara2016generative,miyato2018spectral}. But here (and next section), even though clipping may help robustness empirically, the non-robust standard GAN is still more brittle, also towards the choice of algorithm (amount of regularization), 
than robust GANs. Extending the theory to include both aspects, model and algorithm, is an interesting future avenue.

\subsection{MNIST}
\begin{figure*}[htp]
\vspace{-0.1in}
 \centering
   \includegraphics[width=5.4in]{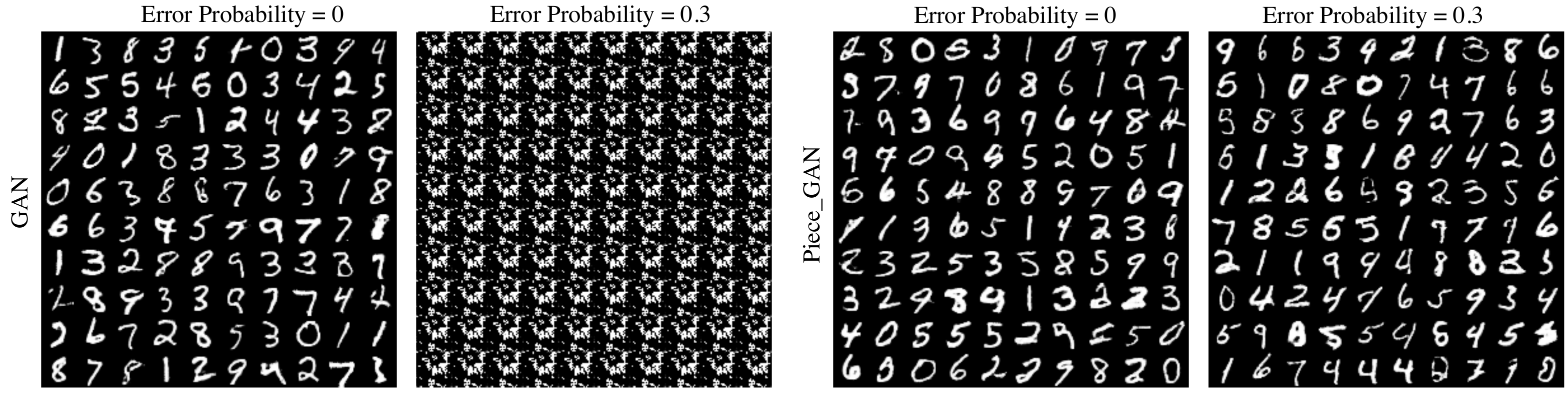}
   \caption{Example results on MNIST (no clipping).}\label{fig:mnist_examples}
   \vspace{-0.1in}
 \end{figure*}
Next, we perform a similar analysis with the MNIST data, using a CNN for both $G$ and $D$. Without clipping, both the GAN and the robust GANs learn the distribution well, and generate all digits with the same probability. Here, we call a learning experiment a success if $G$ learns to generate all digits with the same probability (see Appendix \ref{app:exp_robustness} for a plot of the total variation distance between the distribution of the learned digits and the uniform distribution).
To further explore our theoretical frameworks, we apply a more sophisticated adversary as follows: $\Phi(D(x))$ equals $1-D(x)$ or $\sqrt{D(x)}$ or ${D(x)}^2$, each with probability 0.1, and otherwise, $\Phi(D(x))=D(x)$ (i.e., honest feedback with probability 0.7).
By Theorem~\ref{thm:stronger_2}, all the robust models we explore here should be robust against such a complex adversary.
Figure~\ref{fig:mnist_examples} visualizes the output of the GAN and the ``piecewise linear'' robust GAN with and without an attack. Clearly, the GAN is heavily affected by the attack -- it appears to learn a point mass that maximizes the KL-divergence between $\mathbb{P}_{\text{data}}$ and $\mathbb{P}_{G}$, well in line with the theoretical analysis. The robust GAN, as expected, still performs well. 

\begin{figure}[htp]
 \centering
           \includegraphics[width=4.65in]{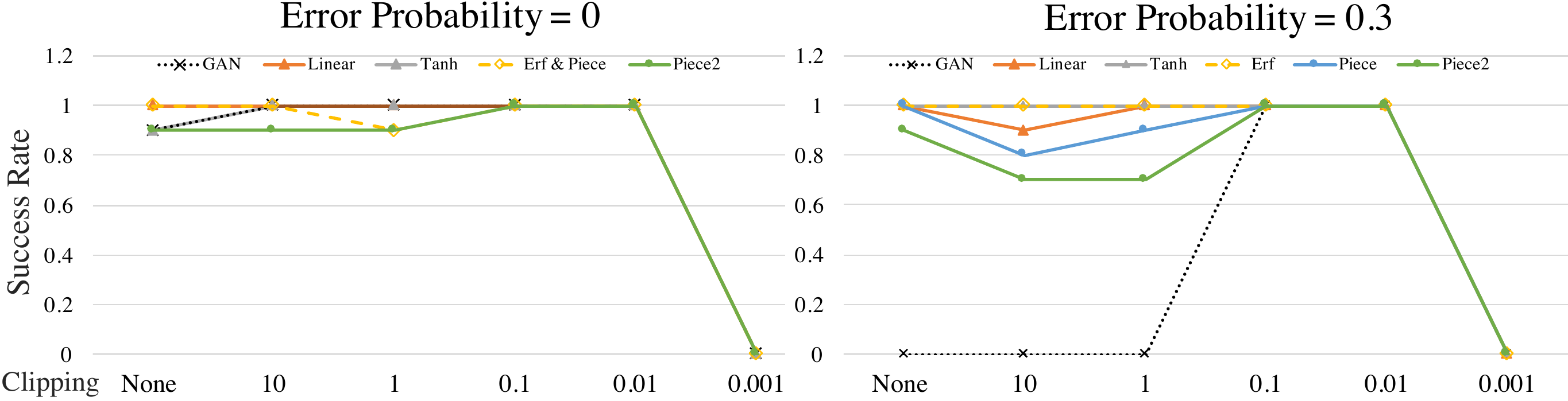}
   \caption{Success rates for various models and clipping values.\label{fig:mnist_success}}
   \vspace{-0.1in}
 \end{figure}

{\bf Clipping.} As for the Gaussians, Figure \ref{fig:mnist_success} shows the success rate over 10 runs for various clipping thresholds, with and without attack. As above, the robust models succeed over a wide range of thresholds, requiring less regularization, both with and without attack. The GAN is stabilized by clipping, but still very sensitive to attacks and fails completely for thresholds above 0.1. That is, over a wide range of regularization parameters, 
the model being robust or not makes a significant difference. 
These results align with our previous observations and support the robust models.

In summary, the experiments demonstrate that the identified robust models are indeed robust against adversaries and have overall more stable training behavior, requiring less regularization in general. 

\section{Empirical Results: Extensions}\label{sec:empirical_edge}

A key property ensuring robustness in Theorems \ref{thm:stronger_1} and \ref{thm:stronger_2} was the symmetry of the transformation functions $f_D$ and $f_G$, i.e., the class $\mathcal{H}$. When deriving the robust models, we took a probabilistic viewpoint, where the outputs of the discriminator is the normalized probability of being the true image (i.e., a sigmoid output layer). However, some recent models, e.g., WGAN, do not apply the sigmoid. Here, we generalize the symmetry property to such raw scores, and obtain a class of models that, as we will see, performs well empirically.


For robust models in Framework 2, i.e.,
$f_D, f_G\in\mathcal{H}$, the GAN objectives (\ref{gen_att_dis}) and (\ref{gen_att_gen}), become
  \begin{align}
    \max_{D}\: \mathbb{E}_{x\sim \bp_{\text{data}}}[f_D( D(x))]-\mathbb{E}_{z\sim \bp_z}[f_D(D(G(z)))] \label{gen_att_dis_model}\\
    \:\min_{G}\:\mathbb{E}_{x\sim \bp_{\text{data}}}[f_G( D(x))] - \mathbb{E}_{z\sim \bp_z}[f_G(D(G(z)))].    \label{gen_att_gen_model}
  \end{align}
  Recall that $\mathcal{H}$ is the set of increasing functions that are also odd functions around 0.5, in the support $[0,1]$, i.e., $\mathcal{H}$ is the set of odd functions around the mid-point of the support. When using raw scores instead of probabilities (i.e., a linear output layer), the support must become $(-\infty,\infty)$. 
%
  Therefore, if we train discriminators using raw scores in (\ref{gen_att_dis_model}) and (\ref{gen_att_gen_model}), a natural, straightforward extension is to consider the set of functions $\hat{\mathcal{H}}$ that is odd around the mid-point of the support $(-\infty,\infty)$, i.e., $0$. Formally, this motivates the following class of functions, $\hat{\mathcal{H}}$: 
\begin{equation*}
\begin{split}
\hat{\mathcal{H}}\triangleq\big\{f(\cdot):f \textrm{ is strictly increasing and differentiable,}
\textrm{ and }f(\theta)=-f(-\theta)\big\}.
\end{split}
\end{equation*}
Hence, when training discriminators with raw scores, like WGAN, we use the objectives (\ref{gen_att_dis_model}) and (\ref{gen_att_gen_model}), and choose functions $f_D\in\hat{\mathcal{H}}$ and $f_G\in\hat{\mathcal{H}}$. When both $f_D$ and $f_G$ are linear functions, this strongly resembles the WGAN.
Due to this analogy, we adopt further specifics used with WGAN: regularizing the Lipschitz constant of the discriminator has been observed to be extremely important \citep{qi2017loss,uehara2016generative,gulrajani2017improved,miyato2018spectral}, so we also add a gradient penalty component, $(||\nabla_{\hat{x}}D(\hat{x})||_2-1)^2$, to the discriminator's objective \citep{gulrajani2017improved}. See Appendix \ref{sec:app:cifar10} for details. In short, this new model class replaces the linear $f$ in WGAN by functions from $\hat{\mathcal{H}}$. 


We test these new models on CIFAR10. By convention, we choose functions such that $f_D=f_G=f$ and $f\in\hat{\mathcal{H}}$. Specifically, we consider the following models: (1) $f(x)= 3\tanh(0.15x)$; (2) $f(x)=5\textrm{erf}(0.1x)$; (3) $f(x) = 2x$ if $x\in[-0.25,0.25]$; $f(x)=\sqrt{x}$ if $x>0.25$; $f(x)=-\sqrt{-x}$ if $x<-0.25$. We add the gradient penalty to the discriminator, and use the WGAN-GP code provided in \cite{gulrajani2017improved}, \emph{without} changing the architecture or hyperparameters to ensure fair comparisons.

\begin{table*}[htp]
\centering
\subfigure{
\begin{tabular}{ cc } 
 \hline
 Method & Inception Score  \\ 
 \hline
 ALI \citep{dumoulin2016adversarially}& $5.34\pm.05$  \\ 
 BEGAN \citep{berthelot2017began} & 5.62\\
 DCGAN \citep{radford2015unsupervised}& $6.16\pm.07$  \\
 Improved GAN (-L+HA) \citep{salimans2016improved} & $6.86\pm.06$\\
 EGAN-Ent-VI \citep{dai2017calibrating} & $7.07\pm.10$\\
DFM \citep{warde2017improving} & $7.72\pm.13$\\
WGAN-GP ResNet \citep{gulrajani2017improved} & $7.86\pm.07$\\
\hline
\hline
(1) Tanh-GP ResNet ({\bf ours}) & $7.80\pm.09$\\
(2) Erf-GP ResNet ({\bf ours}) & $7.76\pm.07$\\
(3) Square-Root-GP ResNet ({\bf ours}) & $7.93\pm.08$\\
 \hline
\end{tabular}}
\hspace{2ex}
\subfigure{
    \vspace{0.5in}
    \includegraphics[width=0.315\textwidth]{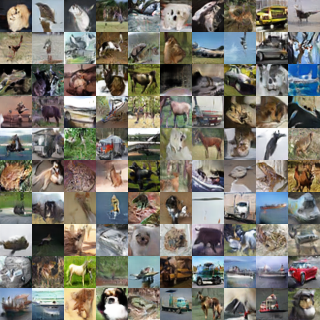} 

}
\vspace{-0.1cm}
\caption{Left: Inception scores on various unsupervised models. The table is borrowed from \cite{gulrajani2017improved} with new entries for our results. Right: Samples from the Square-Root-GP model.}\label{table:inception}
\vspace{-0.1cm}
\end{table*}

Table \ref{table:inception} summarizes inception scores for various models. Our new models yield results competitive with WGAN-GP, without specific tuning. 
Additional figures on learning curves and samples may be found in Appendix \ref{app:cifar_experimental_setup}. 
Overall, these empirical results demonstrates that this new model class arising from our robust class leads to practically appealing results.




\section{Conclusion}
Since the advent of GANs, much effort has been devoted to improving the original formulation. 
In this work, we offer a new viewpoint inspired by the interactive learning dynamics, via probing the robustness of GANs to internal perturbations. 
%
In particular, we identify conditions for the objective function that induce robustness and improve stability more generally.
Inspired by our theoretical results, we further extend our framework to obtain models with promising performance. 
Overall, our study leads to a formal notion of robustness, and opens avenues for further theoretical and empirical analyses of how model parameters, regularization, algorithms and data affect this robustness.

\bibliography{refer}
\bibliographystyle{plain}

%

\appendix
\begin{appendix}
\onecolumn
\begin{center}
\textbf{{\LARGE Supplementary Material}}
\end{center}
\section{Proofs}
This section provides all the missing proofs in the main paper. For convenience, we also repeat the theorems here.
\subsection{Proof of Lemma \ref{claim:vanilla_gan}}
{\bf Lemma \ref{claim:vanilla_gan}.}
Given the optimal discriminator $D^*_G$, the minimization of the objective (\ref{vanilla_g_dishonest}) becomes
\begin{align*}
\min_{G}\quad & 2 \times JSD(\bp_{\text{data}}||\bp_G)-\log(4) -p \left\{KL(\bp_{\text{data}}||\bp_G) + KL(\bp_G||\bp_{\text{data}})\right\}.
\end{align*}
Furthermore, for every $p>0$, the optimal $\bp_G$ can be arbitrarily far from $\bp_{\text{data}}$ in terms of KL-divergence.
\begin{proof}
Given the optimal discriminator $D^*_G(x)=\frac{\bp_{data}(x)}{\bp_{data}(x)+\bp_g(x)}$, the generator's objective (i.e., Eq. (\ref{vanilla_g_dishonest})) becomes
\begin{equation}\label{supp:vanila_g_optimal}
\begin{split}
(\ref{vanilla_g_dishonest}) = \min_{G}\quad &(1-p)\mathbb{E}_{x\sim \bp_{data}}\left[\log \frac{\bp_{data}(x)}{\bp_{data}(x)+\bp_G(x)}\right]+(1-p)\mathbb{E}_{x\sim \bp_G(x)}\left[\log \frac{\bp_{G}(x)}{\bp_{data}(x)+\bp_G(x)}\right]\\
& +p\mathbb{E}_{x\sim \bp_{data}}\left[\log \frac{\bp_{G}(x)}{\bp_{data}(x)+\bp_G(x)}\right]+p\mathbb{E}_{x\sim \bp_G(x)}\left[\log \frac{\bp_{data}(x)}{\bp_{data}(x)+\bp_G(x)}\right]\\
=\min_{G}\quad &\mathbb{E}_{x\sim \bp_{data}}\left[\log \frac{\bp_{data}(x)}{\bp_{data}(x)+\bp_G(x)}\right]+\mathbb{E}_{x\sim \bp_G(x)}\left[\log \frac{\bp_{G}(x)}{\bp_{data}(x)+\bp_G(x)}\right]\\
& +p\mathbb{E}_{x\sim \bp_{data}}\left[\log \frac{\bp_{G}(x)}{\bp_{data}(x)}\right]+p\mathbb{E}_{x\sim \bp_G(x)}\left[\log \frac{\bp_{data}(x)}{\bp_G(x)}\right]\\
=\min_{G}\quad& 2 JSD(\bp_{data}||\bp_{G})-\log4
-p KL(\bp_{data}||\bp_G)-p KL(\bp_{G}||\bp_{data}).
\end{split}
\end{equation}
From the last equality, note that the Jensen-Shannon divergence is bounded, and hence the minimum value is $-\infty$ whenever the error probability $p>0$, which can be achieved, for example, by any $\bp_G$ that concentrates on a particular point. In contrast, if $\bp_G=\bp_{data}$, then the objective achieves value $-\log 4$. Therefore, the discriminator can learn a distribution that is significantly different from $\bp_{data}$.
\end{proof}
\subsection{Proof of Lemma \ref{lemma2}}
{\bf Lemma \ref{lemma2}.} Suppose that $f_D\in\mathcal{H}$, then for a fixed $G$, the optimal $D$ that maximize Eq. (\ref{gen_att_dis}) is
\begin{equation*}
D^*_G(x)=\left\{ \begin{array}{rcl}
1, & \mbox{if }
& \bp_{data}(x)>\bp_g(x), \\ 0, & \mbox{if } & \bp_{data}(x)<\bp_g(x) \\
\emph{$[0,1]$}, & \mbox{if} 
& \bp_{data}(x)=\bp_g(x),
\end{array}\right.
\end{equation*} 
where the notation $D^*_G(x)=[0,1]$ means that $D^*_G(x)$ can be any scalar in the interval $[0,1]$.
\begin{proof} Since $f_D\in\mathcal{H}$, we have 
\begin{equation}\label{supp:eqn_proof_lem2}
\begin{split}
&\mathbb{E}_{x\sim \bp_{data}}[f_D( D(x))]+\mathbb{E}_{z\sim \bp_Z}[f_D(1-D(G(z)))]\\
=&\int f_D(D(x))\bp_{data}(x)+f_D(1-D(x))\bp_G(x)dx\\
=&\int f_D(D(x))(\bp_{data}(x)-\bp_G(x))dx
\end{split}
\end{equation}
Note that $f_D$ is strictly increasing in $[0,1]$ and $D(x)\in[0,1]$. Therefore, when $\bp_{data}(x)>\bp_G(x)$, $f_D(D(x))(\bp_{data}(x)-\bp_G(x))$ is maximized at $D(x)=1$; when $\bp_{data}(x)<\bp_G(x)$, $f_D(D(x))(\bp_{data}(x)-\bp_G(x))$ is maximized at $D(x)=0$. This shows that the integration (\ref{supp:eqn_proof_lem2}) is maximized by the discriminator given in Lemma \ref{lemma2}. 
\end{proof}
\subsection{Proof of Theorem \ref{thm:stronger_1}} \label{supp:theorem1}
{\bf Theorem \ref{thm:stronger_1}.} Suppose that $f_D(\cdot)=\log(\cdot)$ and $f_G\in\mathcal{H}$. Let $\Psi$ be the set of perturbations $\psi:[0,1]\rightarrow[0,1]$ that satisfy {\bf either} one of the following:
\begin{enumerate}
  \item $\psi(\theta)$ is non-decreasing in $[0,1]$ and $\psi(\frac{1}{2})=\frac{1}{2}$;
  \item $\psi(\theta)$ is non-increasing in $[0,1]$, $\psi(\frac{1}{2})=\frac{1}{2}$, and
  \begin{equation*}
\left\{ \begin{array}{rcl}
\psi(\theta)+\theta \geq 1, & \mbox{for }
& \theta\in(\frac{1}{2},1], \\ 
\psi(\theta)+\theta \leq 1, & \mbox{for }
& \theta\in[0,\frac{1}{2}).
\end{array}\right.
\end{equation*}
\end{enumerate}
Then, for any mostly honest adversary $\Phi$ with respect to $\Psi$, given the optimal discriminator, the optimal generator $G^*$ satisfies $\bp_{G^*}(x)=\bp_{data}(x)$. 
\begin{proof}
Fix a mostly honest attack $\Phi$ with respect to the set $\Psi$ defined in Theorem \ref{thm:stronger_1}. By definition, a mostly honest adversary assigns more than 0.5 probability on the function $\psi(\theta)=\theta$. Without loss of generality, denote by $\psi_1$ the previous function, i.e., $\psi_1(\theta)=\theta$. Then, with our notation in Definition \ref{def:attack}, $p_1>0.5$.

Since $f_G\in\mathcal{H}$, we can rewrite the generator's objective function (i.e., Eq.(\ref{gen_att_gen3})) as follows:
\begin{equation}\label{supp:eqn:proof_thm3}
\begin{split}
V&\triangleq \sum_{i=1}^Lp_i\Big(\mathbb{E}_{x\sim \bp_{data}}\Big[f_G\big(\psi_i(D(x))\big)\Big]+\mathbb{E}_{z\sim \bp_Z}\Big[f_G\big(1-\psi_i(D(G(z)))\big)\Big]\Big)\\
&=\sum_{i=1}^Lp_i\Big(\mathbb{E}_{x\sim \bp_{data}}\Big[f_G\big(\psi_i(D(x))\big)\Big]-\mathbb{E}_{x\sim \bp_G}\Big[f_G\big(\psi_i(D(x))\big)\Big]\Big)\\
&=\Big(p_1-\sum_{i=2}^Lp_i\Big)\Big(\mathbb{E}_{x\sim \bp_{data}}\Big[f_G\big(\psi_1(D(x))\big)\Big]-\mathbb{E}_{x\sim \bp_G}\Big[f_G\big(\psi_1(D(x))\big)\Big]\Big)\\
&+\sum_{i=2}^Lp_i\Big(\mathbb{E}_{x\sim \bp_{data}}\Big[f_G\big(\psi_i(D(x))\big)+f_G\big(\psi_1(D(x))\big)\Big]-\mathbb{E}_{x\sim \bp_G}\Big[f_G\big(\psi_i(D(x))\big)+f_G\big(\psi_1(D(x))\big)\Big]\Big)\\
&=\underbrace{\Big(p_1-\sum_{i=2}^Lp_i\Big)\Big(\mathbb{E}_{x\sim \bp_{data}}\Big[f_G\big(D(x)\big)\Big]-\mathbb{E}_{x\sim \bp_G}\Big[f_G\big(D(x)\big)\Big]\Big)}_{V_1}\\
&+\underbrace{\sum_{i=2}^Lp_i\Big(\mathbb{E}_{x\sim \bp_{data}}\Big[f_G\big(\psi_i(D(x))\big)+f_G\big(D(x)\big)\Big]-\mathbb{E}_{x\sim \bp_G}\Big[f_G\big(\psi_i(D(x))\big)+f_G\big(D(x)\big)\Big]\Big)}_{V_2}.
\end{split}
\end{equation}

When $\bp_{data}=\bp_G$, it is obvious that $V_1=V_2=0$. In what follows, we prove the following two facts:
\begin{enumerate}
  \item If $\bp_{data}\neq \bp_G$, then $V_1>0$.
  \item If $\bp_{data}\neq \bp_G$, then $V_2\geq0$.
\end{enumerate}
Combining the two facts, it is clear that in order to minimize $V$, the optimal generator $G^*$ must satisfies $\bp_{data}=\bp_{G^*}$, and this completes the proof of Theorem \ref{thm:stronger_1}.

\emph{Proof of Fact 1:} Note that since $f_D(\cdot)=\log(\cdot)$, the optimal discriminator for a fixed generator $G$ is given by $D(x)=\frac{\bp_{data}(x)}{\bp_{data}(x)+\bp_G(x)}$ \cite{goodfellow2014generative}. With this optimal discriminator, we then have 
\begin{equation*}\label{supp:proof_2}
\begin{split}
\hat{V}_1&\triangleq \mathbb{E}_{x\sim \bp_{data}}[f_G(D(x))]-\mathbb{E}_{x\sim \bp_G}[f_G(D(x))]\\
&=\int \Big\{f_G(D(x))\bp_{data}(x)-f_G(D(x))\bp_G(x)\Big\}dx\\
&=\int f_G\left(\frac{\bp_{data}(x)}{\bp_{data}(x)+\bp_G(x)}\right)(\bp_{data}(x)-\bp_G(x))dx.
\end{split}
\end{equation*}
To show that $\bp_G\neq \bp_{data}$ implies $V>0$, we note that $f_G\in\mathcal{H}$ implies that 
\begin{equation*}
\left\{ \begin{array}{rcl}
f(\theta) >0, & \mbox{for }
& \theta\in(\frac{1}{2},1], \\ f(\frac{1}{2})=0, &  &  \\
f(\theta) <0, & \mbox{for }
& \theta\in[0,\frac{1}{2}).
\end{array}\right.
\end{equation*} 
Therefore, for any $x$ such that $\bp_G(x)\neq \bp_{data}(x)$, we have
\begin{equation*}
f\left(\frac{\bp_{data}(x)}{\bp_{data}(x)+\bp_G(x)}\right)(\bp_{data}(x)-\bp_G(x))>0.
\end{equation*}
This means that $\hat{V}_1>0$ if $\bp_G\neq \bp_{data}$. By assumption, $p_1>0.5$ and hence $p_1-\sum_{i=2}^Lp_i>0$. This completes the proof. Therefore, $V_1=(p_1-\sum_{i=2}^Lp_i)\hat{V}_1>0$ if $\bp_G\neq \bp_{data}$.

\emph{Proof of Fact 2:} Note that
\begin{equation*}\label{supp:eqn:thm4_proof2}
\begin{split}
&\mathbb{E}_{x\sim \bp_{data}}\Big[f_G\big(\psi_i(D(x))\big)+f_G\big(D(x)\big)\Big]-\mathbb{E}_{x\sim \bp_G(x)}\Big[f\big(\psi_i(D(x))\big)+f_G\big(D(x)\big)\Big]\\
&=\int \Big[f_G\big(\psi_i(D(x))\big)+f_G\big(D(x)\big)\Big]\Big(\bp_{data}(x)-\bp_G(x)\Big)dx.
\end{split}
\end{equation*}
For any $x$ such that $\bp_{data}(x)>\bp_G(x)$, the optimal discriminator $D(x)=\frac{\bp_{data}(x)}{\bp_{data}(x)+\bp_G(x)}>\frac{1}{2}$. We then claim that $f_G\big(\psi_i(D(x))\big)+f_G\big(D(x)\big)\geq0$. To see why this must hold, consider first the case where $\psi_i$ satisfies the first condition in Theorem \ref{thm:stronger_1}. Then,
\begin{equation*}
f_G\big(\psi_i(D(x))\big)+f_G\big(D(x)\big)>f\big(\frac{1}{2}\big)+f\big(\frac{1}{2}\big)=0,
\end{equation*}
where the first inequality holds because $f_G$ is strictly increasing and $\psi_i$ is non-decreasing. For the case where $\psi$ satisfies the second condition in Theorem \ref{thm:stronger_1}, since $D(x)>\frac{1}{2}$, we then have $\psi_i(D(x))+D(x)\geq 1$. Therefore, 
\begin{equation*}
f_G\big(\psi_i(D(x))\big)+f_G\big(D(x)\big)\geq f_G\big(1-D(x)\big)+f_G\big(D(x)\big)=-f_G\big(D(x)\big)+f_G\big(D(x)\big)=0,
\end{equation*}
where we have used the property that $f_G(1-\theta)=-f_G(\theta)$.

Similarly, for any $x$ such that $\bp_{data}(x)<\bp_G(x)$, the optimal discriminator $D(x)<\frac{1}{2}$ and we claim that $f_G\big(\psi_i(D(x))\big)+f_G\big(D(x)\big)\leq0$. ff
Consider first the case where $\psi_i$ satisfies the first condition in Theorem \ref{thm:stronger_1}. Then,
\begin{equation*}
f_G\big(\psi_i(D(x))\big)+f_G\big(D(x)\big)<f_G\big(\frac{1}{2}\big)+f_G\big(\frac{1}{2}\big)=0.
\end{equation*}
For the case where $\psi_i$ satisfies the second condition in Theorem \ref{thm:stronger_1}, since $D(x)<\frac{1}{2}$, we then have $\psi_i(D(x))+D(x)\leq 1$ and hence
\begin{equation*}
f_G\big(\psi_i(D(x))\big)+f_G\big(D(x)\big)\leq f_G\big(1-D(x)\big)+f_G\big(D(x)\big)=0.
\end{equation*}

In conclusion, $\Big[f_G\big(\psi_i(D(x))\big)+f_G\big(D(x)\big)\Big]\Big(\bp_{data}(x)-\bp_G(x)\Big)\geq0$ for any $x$ such that $P_{data}(x)\neq P_g(x)$. Therefore, $V_2\geq 0$ and this completes the proof of Fact 2. 
\end{proof}
\subsection{Proof of Theorem \ref{thm:stronger_2}} \label{supp:theorem2}
{\bf Theorem \ref{thm:stronger_2}.} Suppose that $f_D\in\mathcal{H}$ and $f_G\in\mathcal{H}$. Let $\Psi$ be the set of all possible perturbations. Then, for any mostly honest adversary $\Phi$ with respect to $\Psi$, given the optimal discriminator,
the optimal generator satisfies $\bp_{G^*}(x)=\bp_{\text{data}}(x)$.
\begin{proof}
The proof is quite similar to the proof of Theorem \ref{thm:stronger_1}. Since $f_G\in\mathcal{H}$, we can again rewrite the generator's objective function (i.e., Eq.(\ref{gen_att_gen3})) as $V_1+V_2$ (i.e., Eq.(\ref{supp:eqn:proof_thm3}). Obviously, when $\bp_{data}=\bp_G$, it is obvious that $V_1=V_2=0$. Note that since $f_D\in\mathcal{H}$, the optimal discriminator for a fixed generator is now given by Lemma \ref{lemma2}. In the sequel, we follow the proof of Theorem \ref{thm:stronger_1} to show the two facts below when given the optimal discriminator:
\begin{enumerate}
  \item If $\bp_{data}\neq \bp_G$, then $V_1>0$.
  \item If $\bp_{data}\neq \bp_G$, then $V_2\geq0$.
\end{enumerate}
The desired result in Theorem \ref{thm:stronger_2} then immediately follows.

\emph{Proof of Fact 1:}
Let
\begin{equation*}
\begin{split}
\hat{V}_1\triangleq \mathbb{E}_{x\sim \bp_{data}}[f_G(D(x))]-\mathbb{E}_{x\sim \bp_G}[f_G(D(x))]=\int f_G\left(D(x)\right)(\bp_{data}(x)-\bp_G(x))dx.
\end{split}
\end{equation*}
Substitute the optimal discriminator in Lemma \ref{lemma2} into $\hat{V}_1$, it can be readily observed that $V>0$ whenever $\bp_{data}\neq\bp_G$. Specifically, for any $x$ such that $\bp_{data}(x)>\bp_G(x)$, we have $f_G(D(x))=f_G(1)>0$; for any $x$ such that $\bp_{data}(x)<\bp_G(x)$, we have $f_G(D(x))=f_G(0)<0$. Hence, $f_G\left(D(x)\right)(\bp_{data}(x)-\bp_G(x))>0$ for any $x$ such that $\bp_{data}(x)\neq\bp_G(x)$. This implies, together with the assumption that $p_1>0.5$, that $V_1=(p_1-\sum_{i=2}^Lp_i)\hat{V}_1>0$ if $\bp_G\neq \bp_{data}$.

\emph{Proof of Fact 2:}
We now shift gears to $V_2$. Note that 
\begin{equation*}
\begin{split}
&\mathbb{E}_{x\sim \bp_{data}}\Big[f_G\big(\psi_i(D(x))\big)+f_G\big(D(x)\big)\Big]-\mathbb{E}_{x\sim \bp_G(x)}\Big[f\big(\psi_i(D(x))\big)+f_G\big(D(x)\big)\Big]\\
&=\int \Big[f_G\big(\psi_i(D(x))\big)+f_G\big(D(x)\big)\Big]\Big(\bp_{data}(x)-\bp_G(x)\Big)dx.
\end{split}
\end{equation*}
For any $x$ such that $\bp_{data}(x)>\bp_G(x)$, the optimal discriminator in Lemma \ref{lemma2} gives $D(x)=1$. Hence,
\begin{equation}\label{supp:eqn:thm3_proof3}
f_G\big(\psi_i(D(x))\big)+f_G\big(D(x)\big)=f_G\big(\psi_i(1)\big)+f_G\big(1\big)=f_G\big(\psi_i(1)\big)-f_G\big(0\big)\geq0,
\end{equation}
where the second equality follows from the property that $f(\theta)=-f(1-\theta)$ and the last inequality holds because $\psi_i(\theta)\geq 0$ for every $\theta\in[0,1]$ and $f_G$ is strictly increasing.

Similarly, for any $x$ such that $\bp_{data}(x)<\bp_G(x)$, the optimal discriminator gives $D(x)=0$ and we then have
\begin{equation}\label{supp:eqn:thm3_proof4}
f\big(\psi_i(D(x))\big)+f\big(D(x)\big)=f\big(\psi_i(0)\big)+f\big(0\big)=f\big(\psi_i(0)\big)-f\big(1\big)\leq0.
\end{equation}

In summary, we have $\Big[f_G\big(\psi_i(D(x))\big)+f_G\big(D(x)\big)\Big]\Big(\bp_{data}(x)-\bp_G(x)\Big)\geq0$ for any $x$ such that $\bp_{data}(x)\neq \bp_G(x)$. 
Therefore, $V_2\geq0$ and this completes the proof of Fact 2.
\end{proof}

\newpage
\section{Experimental Details: Verify Robustness}\label{app:exp_robustness}
In this section, we show the details of our experiments as well as figures omitted from the main text. {While the theory assumes the optimal discriminator, practical training relies on gradient-based algorithms. In our experiments, with the presence of an adversary, each step of training the generator consists of a forward pass, where the generated images pass through the discriminator $D$ and then the adversary $\Phi$ to produce the signal $\Phi(D(x))$, and a backward pass, where the gradients the generator received are computed by backpropagating through the adversary $\Phi$ first and then the discriminator $D$. The presence of an adversary affects the signals as well as the training gradients the generator received. From the viewpoint of the generator, the discriminator and the adversary as a whole can be viewed as a ``dishonest discriminator" that, upon receiving the genrated images, produces a noisy signal and the corresponding gradients for the generator. For all the experiments, we train the models by alternating between updating the generator and the discriminator. If the experiment involves clipping, then a full update step consists of first clipping the weights of the discriminator, and then updating the discriminator and the generator once.}
\subsection{Mixture of Gaussians}\label{sec:Supp:gaussian}
The synthetic data is generated from a mixture of 8 two-dimensional Gaussians with equal variance but different means evenly spaced on a circle. We fix the network architectures and hyper-parameters throughout the experiments. While it is possible to boost the individual performance by adapting the hyper parameters to different models and error probabilities, our focus in this section is to establish a fair comparison among different models and dishonest adversaries.  

The generator consists of a fully connected network with 3 hidden layers, each of size 64 with ReLU activations. The output layer contains two neurons that linearly project the input to 2 dimensions. The discriminator consists of a fully connected network with 3 hidden layers, each of size 256 with ReLU activations, followed by a sigmoid output layer. The latent vectors are sampled from a 256-dimensional multivariate Gaussian distribution with 0 mean and identity covariance matrix. 
For training algorithms, we use Adam with a learning rate of $1\mathrm{E-}4$ and $\beta_1=0.5$ for the generator and RMSprop with a learning rate of $1\mathrm{E-}4$ for the discriminator. The size of each minibatch is fixed to 512. Finally, all the models are trained for 50k steps, 100k steps, and 180k steps when the error probabilities are 0, 0.2, and 0.4, respectively.
\subsubsection{Addition Discussion on Empirical Robustness via Regularization}\label{supp:gaussian_dis}
In our main text (cf. Section \ref{sec:exp_gaussian}), we have argued the two different robustness mechanisms: the objective function we developed and the empirical robustness via regularization. In particular, while the standard GAN formulation is not robust, if the regularization is strong enough (e.g. 0.05 clipping threshold), it may help regularize and improve the empirical stability. The two mechanisms are orthogonal and we now provide evidence to show their difference. This will help to confirm that there is no contradiction to our theory, i.e., the standard GAN \emph{formulation} is not robust.

\begin{figure}[htp]
 \centering
   \includegraphics[width=5.4in]{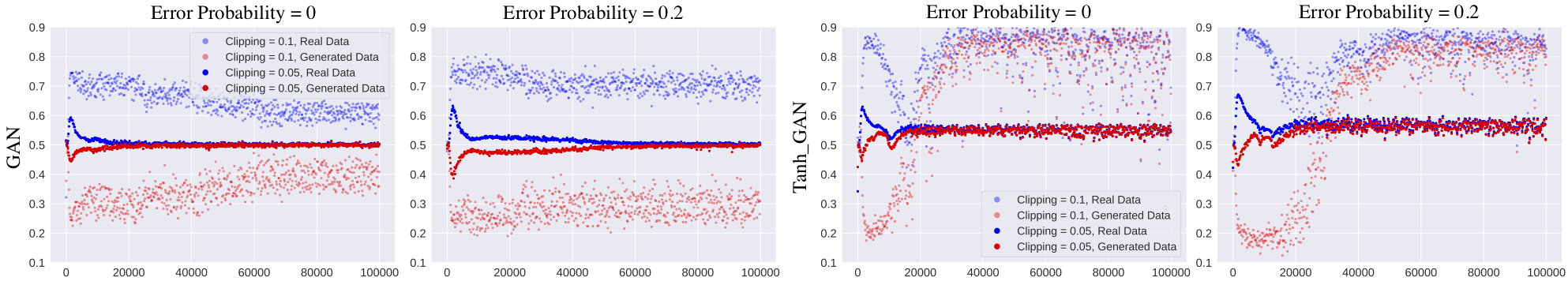}
   \caption{Discriminator's outputs averaged over each batch of data. The blue points represent the outputs for the true data, while the red points represent the outputs for the generated data.}\label{fig:guassian_gan_005} 
 \end{figure}

To this end, we show in Figure \ref{fig:guassian_gan_005} the probabilities $D$ assigns to the true and the generated data (i.e., $D$'s output before applying any dishonest adversary) when the clipping threshold is either $0.1$ or $0.05$. Note the obvious difference between strong regularization (clipping at 0.05) and the weaker one (clipping at 0.1). For strong regularization, the empirical robustness via regularization mechanism is dominant: $D$'s outputs are restricted to be around 0.5\footnote{Due to small clipping threshold, $D$'s weights are restricted to be so small that the unnormalized logits concentrate around $0$ (i.e., 0.5 after passing through the sigmoid layer).}, for both real and the generated data, and hence, regularization prevents extreme outputs that generally impair training. In contrast, with less regularization, $D$'s weights have more freedom (i.e., $D$'s outputs are not restricted to 0.5), and the other mechanism, robust objective functions, now becomes the dominant one: the robust models learns the data distribution consistently, with or without the adversary, while the unrobust standard GAN fails to do so. 


Before closing, we point out one more interesting observation from the robust model (left two plots in Figure~\ref{fig:guassian_gan_005}) that matches our theory. Recall that if $\bp_{\text{data}}=\bp_G$, the actual output value of the optimal discriminator in Lemma~\ref{lemma2} can be any value in $[0,1]$. Indeed, we observe this behavior for the robust model: for large clipping threshold, $D$'s output converges to be around 0.85 instead of 0.5.



\subsubsection{Additional Samples}
Here, we collect all the results for the robust models that are omitted in Figure \ref{fig:gaussian_example}. Those models consistently learn the mixture distribution with or without an adversary.
\begin{figure}[htp]
 \centering
   \includegraphics[width=5.4in]{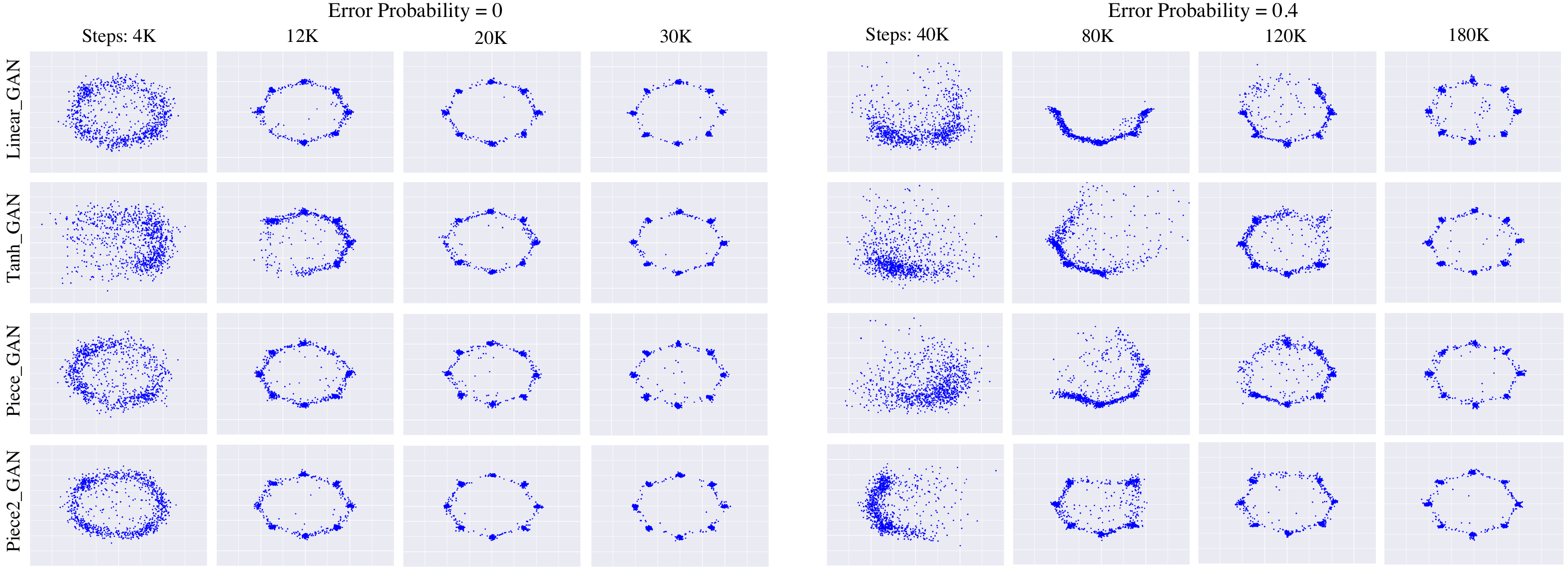}
   \caption{Additional samples that are omitted in Figure \ref{fig:gaussian_example}.}\label{fig:supp:gaussian_sample}
   \vspace{-0.2in}
 \end{figure}
\subsubsection{Averaged Number of Learned Modes}
\begin{figure}[htp]
 \centering
   \includegraphics[width=5.4in]{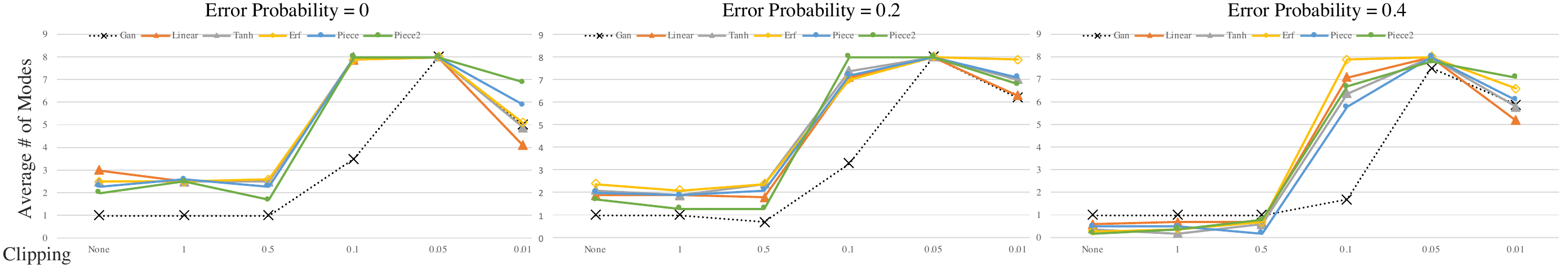}
   \caption{Averaged number of modes learned by each model.}\label{fig:supp:gaussian_mode}
 \end{figure}

Figure \ref{fig:supp:gaussian_mode} supplements the results presented in Figure \ref{fig:guassian_1}. Recall that for each model and each parameter setting, we run 10 experiments. Figure \ref{fig:supp:gaussian_mode} shows the averaged number of modes learned by each model. The results are consistent with what we presented in the main text (cf., Figure \ref{fig:guassian_1} and the corresponding discussion), namely, the robust models tend to perform better under various settings.  
\subsubsection{Averaged Number of Steps for a Successful Learning}
\begin{figure}[htp]
 \centering
   \includegraphics[width=5.4in]{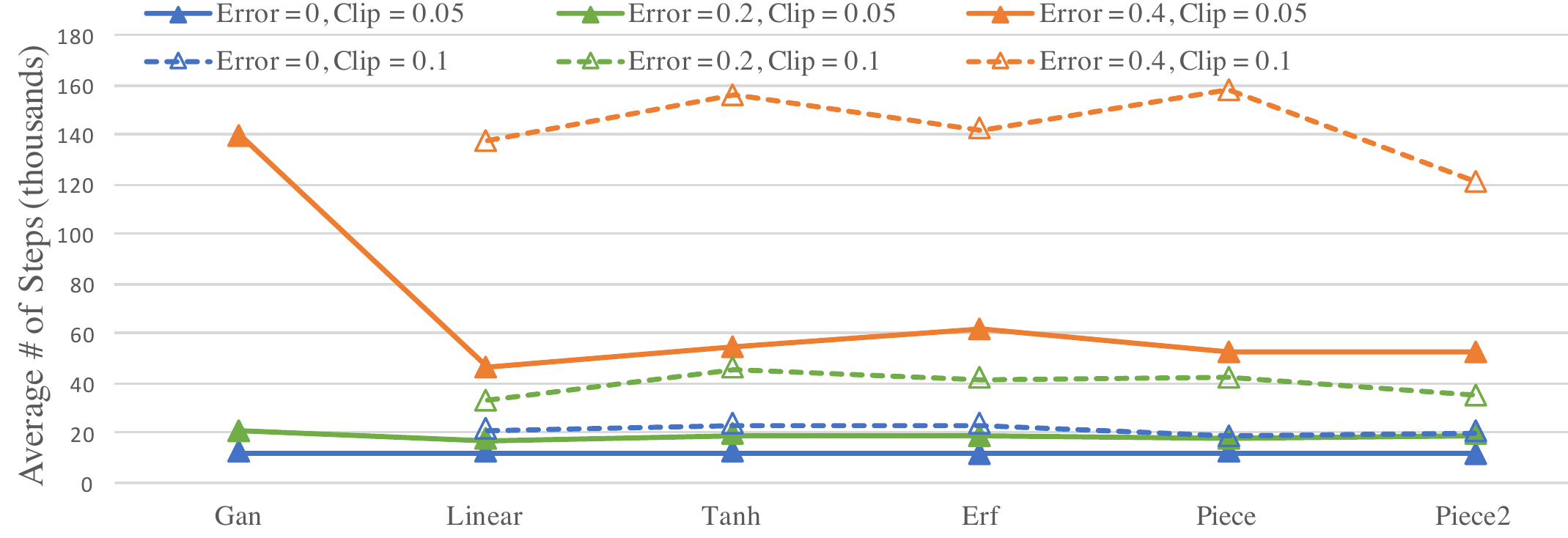}
   \caption{Averaged number of steps for a successful learning. Recall that a success means learning all th 8 modes.}\label{fig:supp:gaussian_step}
 \end{figure}

For each experiment, if the model successfully learns all the 8 modes, we count the number of steps needed and report the average steps in Figure \ref{fig:supp:gaussian_step}. This confirms our intuition: the larger the error probability is, the more steps the robust models will need to average out the noise and extract the right signal to help the overall learning.
\subsection{MNIST}\label{supp:sec:mnist}
We fix the network architectures and hyper-parameters throughout the experiments. The network is adapted from a publicly available CNN model\footnote{\url{https://github.com/hwalsuklee/tensorflow-generative-model-collections}}. In particular, we remove the Batch Normalization layers in the generator. The reason for this is to minimize the effect of architectures on robustness so that we can control as many factors as possible and fairly evaluate how the model \emph{itself} affects the overall robustness. On the other hand, BN is kept for the discriminator. Our theory relies on ideal assumptions of an optimal discriminator. Hence, to verify the theory, it would be beneficial to have a nice discriminator that can discriminate the true and generated data, and provide useful signals.  

We alternate between updating the generator and the discriminator with a minibatch of size 100. The latent vectors are sampled from the uniform distribution on $[-1,1]^{256}$. The generator is trained using Adam with a learning rate of $2\mathrm{E-}4$ and $\beta_1=0.5$, while the discriminator is trained by RMSprop with a learning rate of $5\mathrm{E-}5$. Each model is trained for 50 epochs.
\subsubsection{Additional Discussion on Empirical Robustness via Regularization}
\begin{figure}[htp]
 \centering
   \includegraphics[width=5.4in]{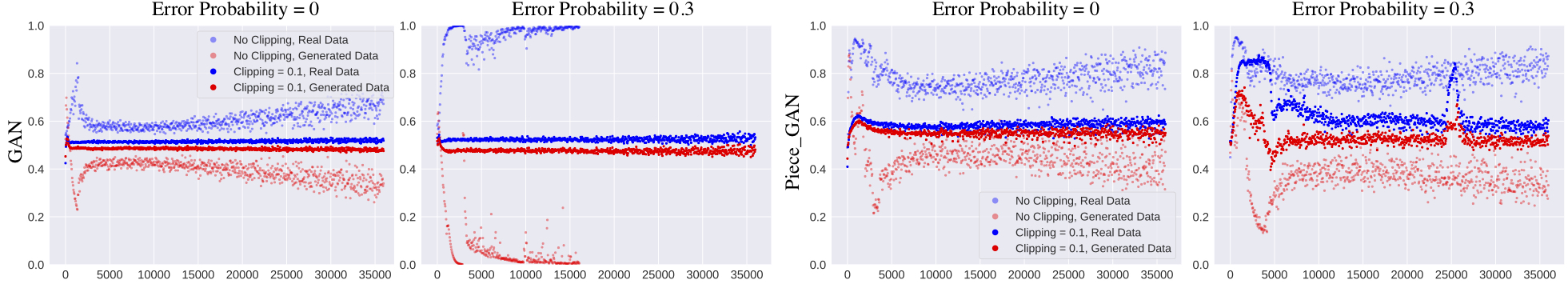}
   \caption{Discriminator's outputs averaged over each batch of data. 
   Upper right: no data points for the lighter plots after 15K steps because $D$'s outputs stay at extreme values for a long time, causing exploding gradients for the generator and eventually numerical errors (nan).}\label{fig:mnist_plot}
   \vspace{-0.1in}
 \end{figure}
This section reinforces our understanding about the two robustness mechanisms, as discussed in Section \ref{supp:gaussian_dis}. Similarly, 
Figure \ref{fig:mnist_plot} shows the discriminator's outputs for the true and generated images, without clipping and with clipping at $0.1$. The figure indicates a similar phenomenon as for the Gaussians. A strong regularization improves empirical robustness by restricting $D$'s outputs from extreme values. When less regularization is applied, the objective function being robust or not becomes a dominant effect: with an adversary, the robust model is still successful while the standard GAN fails completely.


\subsubsection{Additional Samples}
We show results for those robust models that are not presented in Figure \ref{fig:mnist_examples}. Again, the robust models is able to defend the adversary and learn to generate the desired digits with no apparent mode collapse.  
\begin{figure}[htp]
 \centering
   \includegraphics[width=5.4in]{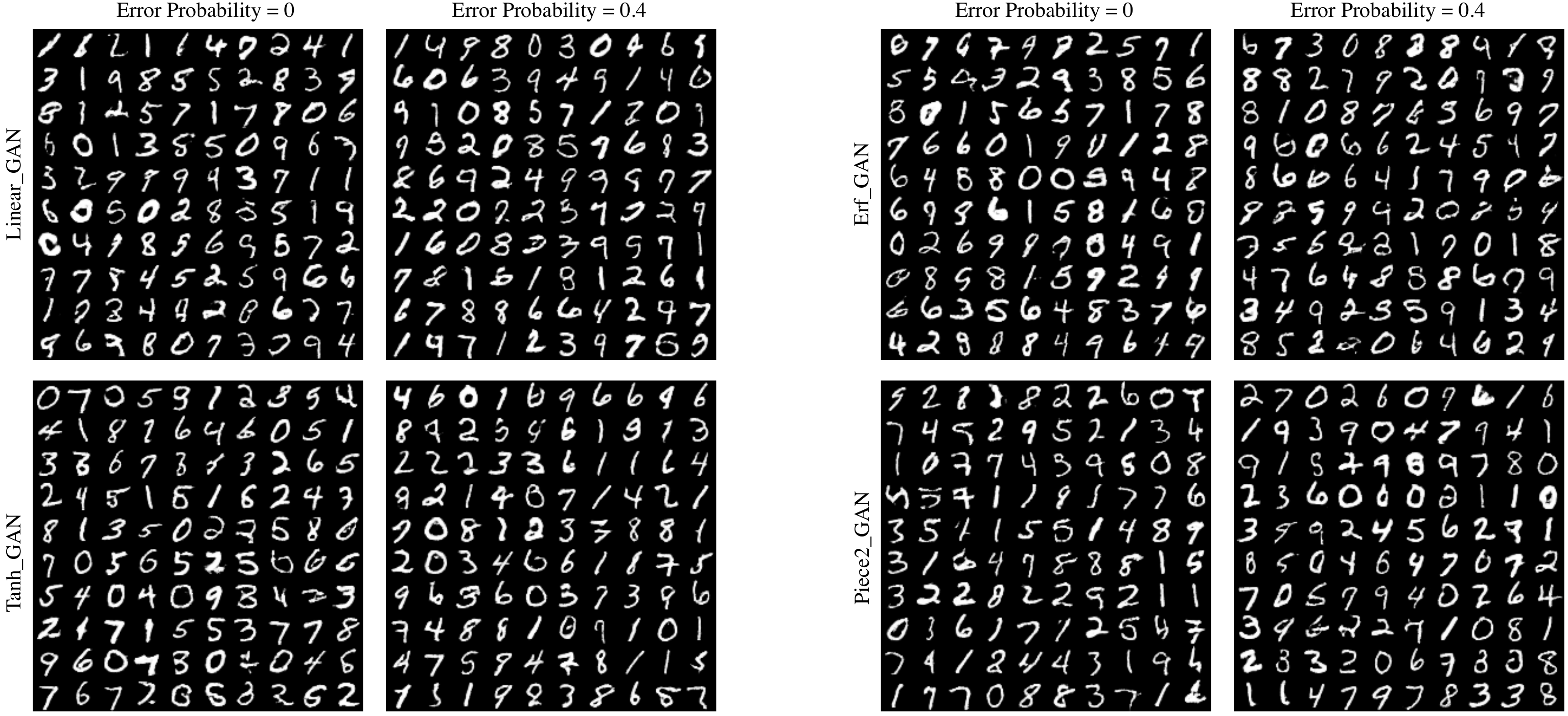}
   \caption{Additional samples that are omitted in Figure \ref{fig:mnist_examples}.}\label{supp:fig:mnist_sample}
\end{figure}
\subsubsection{TV Distance to the Uniform Distribution}
\begin{figure}[htp]
 \centering
   \includegraphics[width=5.4in]{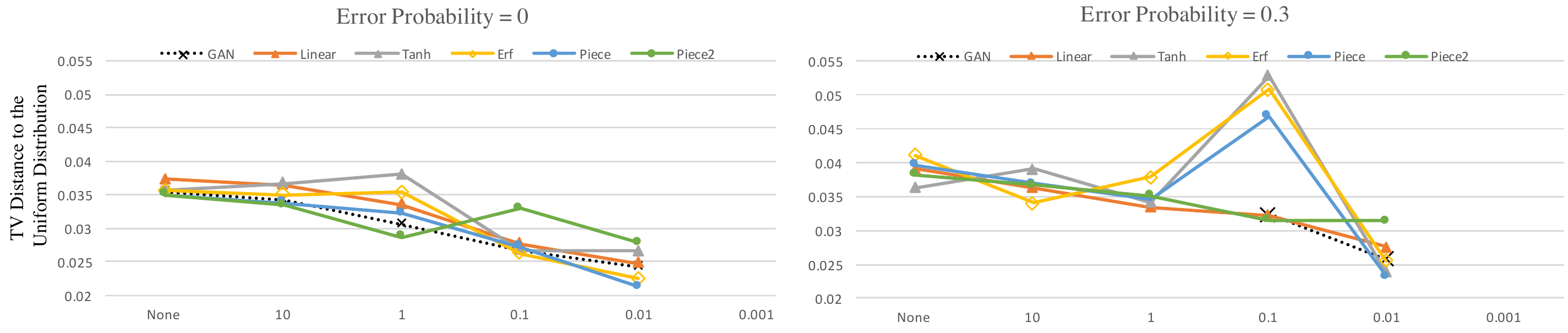}
   \caption{TV distance to the uniform distribution, averaged over \emph{successful} runs.}\label{supp:fig:mnist_tv}
\end{figure}

Figure \ref{supp:fig:mnist_tv} supplements the plots of success rates in Figure \ref{fig:mnist_success}. For each experiment, when the model learns to generate digits, we use an auxiliary classifier to classify the generated data and compute the total variation distance between the learned distribution to the uniform distribution over the 10 digits. Recall that for each model and each parameter setting, we independently run the experiments for 10 times. Figure \ref{supp:fig:mnist_tv} shows the resulting total variation distance, averaged over \emph{successful} runs. Note that the distance is uniformly small, implying that for the successful runs, there is no mode collapse. Consequently, this justifies that we can focus on the plot of success rate in Figure \ref{fig:mnist_success} and draw conclusions correspondingly.

\section{Experimental Details: Extensions}\label{sec:app:cifar10}
For convenience, we restate the models, extended from our theoretical results. The objectives are 
\begin{align}
\label{eqn:app_dis}
    \max_{D}\: \mathbb{E}_{x\sim \bp_{\text{data}}}[f_D( D(x))]-\mathbb{E}_{z\sim \bp_z}[f_D(D(G(z)))],\\
    \min_{G}\:\mathbb{E}_{x\sim \bp_{\text{data}}}[f_G( D(x))] - \mathbb{E}_{z\sim \bp_z}[f_G(D(G(z)))],
  \end{align}
 where $f_D\in\hat{\mathcal{H}}$ and $f_G\in\hat{\mathcal{H}}$. Recall that $\hat{\mathcal{H}}$ are the set of strictly increasing functions that are also odd functions around 0. The output of the discriminator is assumed to be the unnormalized raw scores, which lies in $(-\infty,+\infty)$, instead of probabilities.
\subsection{Gradient Penalty}
Since the model that uses the linear functions in $\hat{\mathcal{H}}$ is almost the same as the Wasserstein GAN, this motives us to use the successful techniques that have been developed for WGAN to train our models. In particular, we observe that gradient penalty can be seamlessly applied in our new framework. We leverage this to also add the same gradient penalty component in WGAN-GP\citep{gulrajani2017improved} to regularize the discriminator. Formally, instead of (\ref{eqn:app_dis}), the discriminator's objective now becomes:
\begin{equation*}
\min_{D}\: -\mathbb{E}_{x\sim \bp_{\text{data}}}[f_D( D(x))]+\mathbb{E}_{z\sim \bp_z}[f_D(D(G(z)))] +\lambda\mathbb{E}_{\hat{x}\sim\bp_{\hat{x}}}[(||\nabla_{\hat{x}}D(\hat{x})-1)^2]
\end{equation*}

For completeness, we restate the WGAN-GP algorithm in \cite{gulrajani2017improved}, and highlight the differences for our models in red. Note that the overall algorithm is exactly as in \cite{gulrajani2017improved}, except for very few places. This means that our models can be easily applied, with almost zero modifications on the existing implementations.

\begin{algorithm}
\caption{Our framework with gradient penalty. Same as in WGAN-GP, we use values of $\lambda=10$, $n_D = 5$, $\alpha=0.0001$, $\beta_1=0$, $\beta_2=0.9$}
\begin{algorithmic}[1]
\REQUIRE the gradient penalty coefficient $\lambda$; the number of discriminator iterations per generator iteration $n_D$; the batch size $m$; Adam hyperparameters $\alpha$, $\beta_1$ and $\beta_2$; the discriminator parameters $w$; the generator parameters $\theta$ 
\WHILE{$\theta$ has not converged}
\FOR{$t=1,\dots,n_D$}
\FOR{$i=1,\dots,m$}
\STATE Sample real data $x\sim \bp_{data}$, latent variable $z\sim\bp_z$, a random number $\epsilon\sim U[0,1]$.
\STATE $\tilde{x}\leftarrow G_\theta(z)$
\STATE $\hat{x}\leftarrow \epsilon x +(1-\epsilon)\tilde{x}$
\STATE $L^{(i)}\leftarrow {\color{red} f_D(}D_w(\tilde{x}){\color{red})} - {\color{red} f_D(}D_w({x}){\color{red})} + \lambda (||\nabla_{\hat{x}}D_w(\hat{x})||_2-1)^2$
\ENDFOR
\STATE $w\leftarrow \textrm{Adam}(\nabla_w\frac{1}{m}\sum_{i=1}^mL^{(i)},w,\alpha,\beta_1,\beta_2)$
\ENDFOR
\STATE Sample a batch of latent variables $\{z^{(i)}\}_{i-1}^m\sim\bp_z$.
\STATE $\theta\leftarrow \textrm{Adam}(\nabla_\theta\frac{1}{m}\sum_{i=1}^m-{\color{red} f_G(}D_w(G_\theta(z)){\color{red})},\theta,\alpha,\beta_1,\beta_2)$
\ENDWHILE 
\end{algorithmic}
\end{algorithm}
\vspace{-0.1in}
\subsection{Experimental Setup}
\label{app:cifar_experimental_setup}
\begin{figure}[h]
\centering
  \centering
  \includegraphics[width=3in]{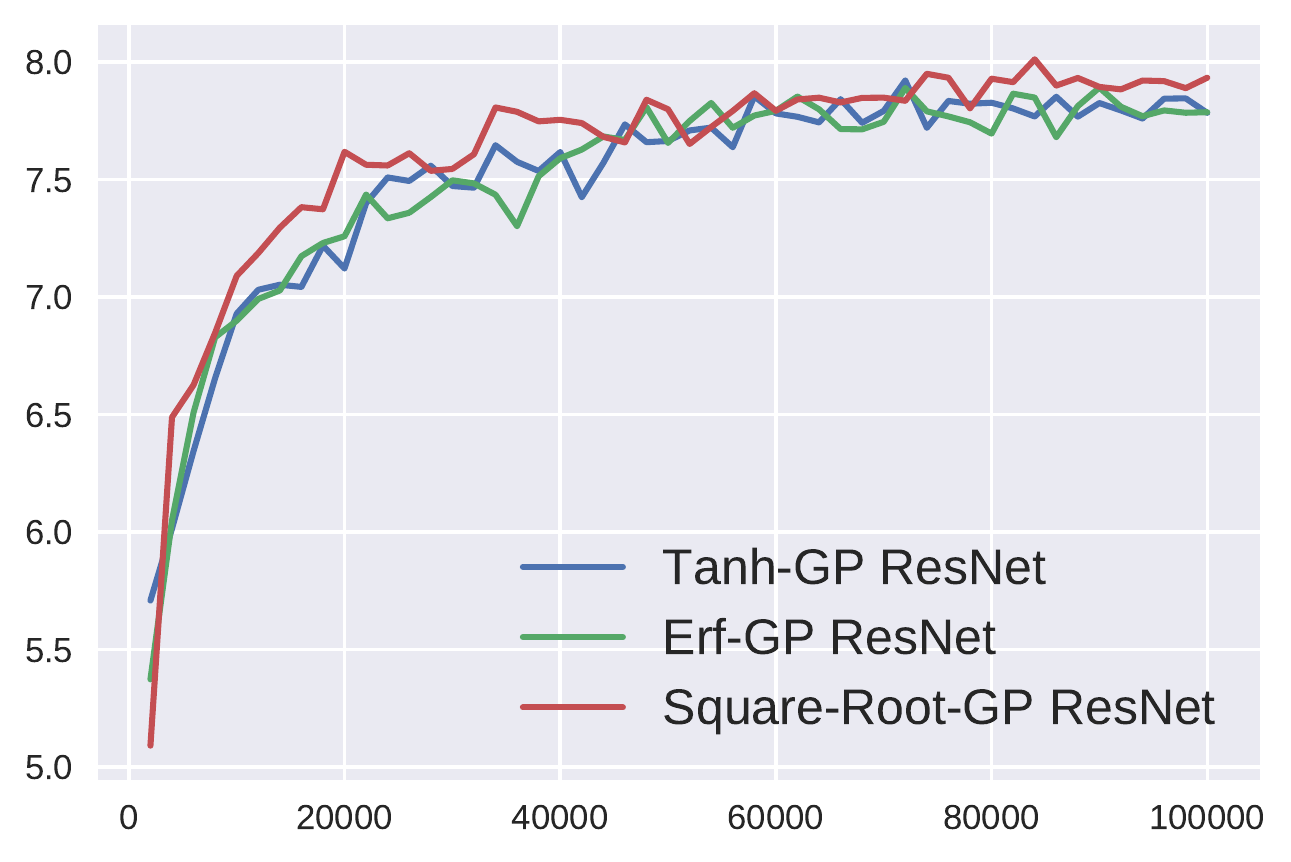}
\caption{Inception Score vs. Iterations}\label{fig:inception}
\end{figure}

We use the publicly available implementation of WGAN-GP with residual network that was used in the original paper \cite{gulrajani2017improved}. \footnote{\url{https://github.com/igul222/improved_wgan_training/blob/master/gan_cifar_resnet.py}} All the hyperparameters and the network architectures are fixed to the original settings in the code, without any modifications. We only modify the loss functions to be those investigated in Section \ref{sec:empirical_edge}. Figure \ref{fig:inception} shows how the inception score increases over iterations. Additional samples that are missing from the main text are shown below.

\begin{figure}[htp]
\begin{minipage}{.5\textwidth}
\centering
\includegraphics[width=.92\textwidth]{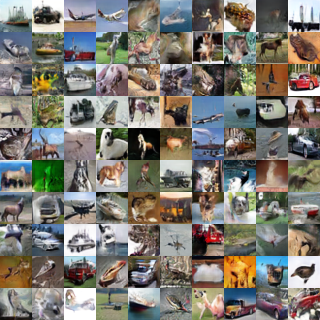}
\caption{Tanh-GP ResNet}
\end{minipage}\hfill
\begin{minipage}{.5\textwidth}
\centering
\includegraphics[width=.92\textwidth]{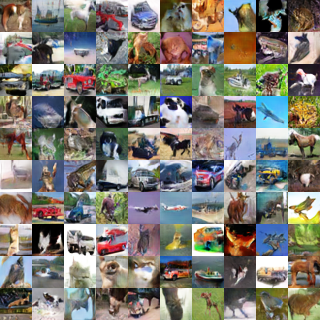}
\caption{Erf-GP ResNet}
\end{minipage}
\end{figure}

\end{appendix}

\end{document}